\documentclass[conference]{IEEEtran}
\IEEEoverridecommandlockouts
\usepackage{cite}
\usepackage{amsmath,amssymb,amsfonts}
\usepackage{algorithmic}
\usepackage{graphicx}
\graphicspath{{./}{sections/}{images/}}
\usepackage{textcomp}
\usepackage[table]{xcolor}
\usepackage{booktabs}
\usepackage{tabularx}
\usepackage{array}
\usepackage{multirow}
\usepackage{xurl}
\usepackage{tikz}
\usetikzlibrary{arrows.meta,positioning,fit,calc}

\newcolumntype{C}[1]{>{\centering\arraybackslash}p{#1}}
\newcolumntype{L}[1]{>{\raggedright\arraybackslash}p{#1}}
\newcolumntype{Y}{>{\raggedright\arraybackslash}X}
\definecolor{perceptionfill}{HTML}{E8F2E3}
\definecolor{decisionfill}{HTML}{F5E4EA}
\definecolor{executionfill}{HTML}{FCEFCF}
\definecolor{stageyellow}{HTML}{F8E9B6}
\definecolor{stageblue}{HTML}{DDEAF7}
\definecolor{stageorange}{HTML}{F9DDD1}
\definecolor{stagegreen}{HTML}{E0F0DE}

\def\BibTeX{{\rm B\kern-.05em{\sc i\kern-.025em b}\kern-.08em
    T\kern-.1667em\lower.7ex\hbox{E}\kern-.125emX}}

\begin{document}

\title{Securing Computer-Use Agents: A Unified Architecture--Lifecycle Framework for Deployment-Grounded Reliability}

\author{
\IEEEauthorblockN{Zejian Chen\IEEEauthorrefmark{1}, Zhanyuan Liu\IEEEauthorrefmark{1}, Chaozhuo Li\IEEEauthorrefmark{1}, Mengxiang Han\IEEEauthorrefmark{2},
Songyang Liu\IEEEauthorrefmark{1},\\ Litian Zhang\IEEEauthorrefmark{1}, Feng Gao\IEEEauthorrefmark{2}, Yiming Hei\IEEEauthorrefmark{3}, Xi Zhang\IEEEauthorrefmark{1}\thanks{Corresponding author: Xi Zhang (zhangx@bupt.edu.cn).}}
\IEEEauthorblockA{\IEEEauthorrefmark{1}Beijing University of Posts and Telecommunications, Beijing, China \\
Emails: \{chenzejian, doubao, lichaozhuo, liusyang, zhanglitian, zhangx\}@bupt.edu.cn}
\IEEEauthorblockA{\IEEEauthorrefmark{2}China Unicom, Beijing, China \\
Emails: hanmx12@chinaunicom.cn, gaofeng149@chinaunicom.cn}
\IEEEauthorblockA{\IEEEauthorrefmark{3}China Academy of Information and Communications Technology, Artificial Intelligence Institute, Beijing, China \\
Email: heiyiming@caict.ac.cn}
}

\maketitle

\begin{abstract}
Computer-use agents (CUAs) are moving from bounded benchmarks toward real software environments, where they operate browsers, desktops, mobile applications, filesystems, terminals, and tool backends. In such settings, reliability is no longer captured by task success alone: perception errors, planning drift, memory use, tool mediation, permission scope, and runtime oversight jointly determine whether agent actions remain aligned with user intent. Existing surveys organize the CUA landscape by methods, platforms, benchmarks, or security threats, but less explicitly connect capability formation, authority exposure, failure manifestation, and control placement. To address this gap, the article develops an architecture--lifecycle framework for deployment-grounded reliability in CUAs. The architectural view analyzes \emph{Perception}, \emph{Decision}, and \emph{Execution} as coupled layers that transform software observations into authority-bearing actions. The lifecycle view examines \emph{Creation}, \emph{Deployment}, \emph{Operation}, and \emph{Maintenance} as stages in which priors are learned, tools and permissions are bound, runtime trajectories are stressed, and assurance must be preserved under drift. Using this lens, the analysis synthesizes representative systems, benchmarks, and security/privacy studies; distinguishes where failures become visible from where their enabling conditions are introduced; and maps recurring intervention surfaces for control, oversight, and assurance. OpenClaw is used only as a public motivating example of an open deployment pattern, not as a verified internal case study. The conclusion highlights open challenges in controllable grounding, long-horizon constraint preservation, safe authority binding, mixed-trust runtime defense, privacy-preserving memory, and continual assurance.
\end{abstract}


\section{Introduction}
\label{sec:intro}

Computer-use agents (CUAs) are increasingly studied as agents that can operate real software environments rather than answer prompts alone. Recent systems already span browsers, desktops, mobile applications, filesystems, terminals, and tool backends~\cite{wang2025opencua,gonzalez2025unreasonable,osu_xu2026_mobile-agent-v3-5-multi-platform,osu_zhou2025_mai-ui-technical-report-real-wor,osu_yan2025_step-gui-technical-report,osu_andreux2025_surfer-2-the-next-generation-of,osu_gupta2026_molmoweb-open-visual-web-agent-a}. This transition changes the meaning of reliability. In a dialogue-only setting, a local mistake often remains textual. In a live software environment, the same mistake can become a deleted file, a leaked secret, an unintended transfer, or a persistent misconfiguration. Reliable computer use is therefore a systems problem as much as a modeling problem, because perception, planning, execution authority, memory, tool use, and oversight interact under live software conditions.

The recent expansion of CUA deployment settings makes an integrative survey timely. Benchmarks have moved from bounded website tasks toward visually grounded, enterprise, personalized, and open-environment settings~\cite{deng2023mind2web,zhou2023webarena,koh2024visualwebarena,drouin2024workarena,osu_boisvert2024_workarena-towards-compositional,osu_yuan2026_webforge-breaking-the-realism-re,osu_zhang2026_clawbench-can-ai-agents-complete,osu_chen2026_knowu-bench-towards-interactive,osu_nie2026_pspa-bench-a-personalized-benchm,xie2024osworld}. At the same time, system-building and evaluation work has diversified across grounding, memory, long-horizon planning, tool-augmented execution, safety evaluation, and open-deployment stacks~\cite{wang2025opencua,gonzalez2025unreasonable,osu_wang2026_colorbrowseragent-complex-long-h,osu_yu2026_graphpilot-gui-task-automation-w,osu_zhang2026_showui-aloha-human-taught-gui-ag}. The difficulty is no longer only the lack of evidence about CUA capability. It is also the lack of a common coordinate system for interpreting how capability is formed, how it is bound to operational authority, where failures first become visible, and where controls can intervene.

Existing surveys clarify important slices of this landscape. Current overviews summarize method families, platform coverage, benchmark inventories, and high-level ecosystem structure~\cite{nguyen2025gui,hu2025agents,tang2025survey,osu_wang2024_gui-agents-with-foundation-model,osu_gao2024_generalist-virtual-agents-a-surv,shi2025trustworthygui}. More focused surveys examine reinforcement-learning enhancement, phone automation, WebAgents, or safety and security threats~\cite{osu_li2025_a-survey-on-gui-agents-with-foun,osu_liu2025_llm-powered-gui-agents-in-phone,osu_ning2025_a-survey-of-webagents-towards-ne,osu_chen2025_a-survey-on-the-safety-and-secur}. These works are valuable, but they usually organize the field by method, platform, benchmark, or threat category. Less explicit is a deployment-grounded account of how learned capabilities become authority-bearing actions, how similar runtime failures can originate from different upstream conditions, and how security, privacy, and oversight mechanisms should be placed across both system architecture and lifecycle stage.

To address that gap, the article develops an analytical framework for deployment-grounded reliability in general-purpose CUAs across web, desktop, mobile, and cross-application settings. The framework combines two views. The architectural view analyzes CUAs through three coupled layers: \emph{Perception}, which reconstructs actionable state from software observations; \emph{Decision}, which preserves task-conditioned intent under uncertainty and long-horizon pressure; and \emph{Execution}, which converts decisions into authority-bearing operations. The lifecycle view analyzes four stages: \emph{Creation}, where priors, grounding habits, objectives, and action abstractions are formed; \emph{Deployment}, where tools, sessions, permissions, and observation channels are bound to the agent; \emph{Operation}, where active trajectories are stressed by mixed-trust inputs, partial observability, and asynchronous change; and \emph{Maintenance}, where models, interfaces, tools, and ecosystems drift after release. Together, these two views provide a diagnostic lens for distinguishing where a failure manifests from where its enabling condition was introduced.

The scope here is deployment-grounded reliability rather than CUA capability alone. The analysis focuses on systems, benchmarks, deployment patterns, and security or privacy studies that materially inform how CUAs behave once they interact with real software surfaces and operational authority. OpenClaw is used only as a public motivating example of an open deployment pattern, not as a verified internal case study~\cite{openclaw2026website,ibm2026openclaw}. More generally, the analysis targets the broader class of CUAs that combine interface interaction, tooling, memory, and authority-bearing execution under real-world deployment conditions.

The resulting contributions are:
\begin{itemize}
  \item \textbf{A deployment-grounded reliability scope for CUAs.} CUA reliability is characterized as a joint property of software observation, task-conditioned decision making, authority-bearing execution, memory, tool use, and oversight, rather than as benchmark task success alone.
  \item \textbf{An architecture--lifecycle framework.} The literature is organized through a tri-layer architecture of \emph{Perception}, \emph{Decision}, and \emph{Execution}, together with a four-stage lifecycle of \emph{Creation}, \emph{Deployment}, \emph{Operation}, and \emph{Maintenance}.
  \item \textbf{A failure-origin analysis.} The analysis distinguishes where failures become visible from where their enabling conditions are introduced, relating grounding errors, planning drift, over-privilege, memory misuse, mixed-trust inputs, privacy leakage, and ecosystem drift within one diagnostic view.
  \item \textbf{A control and governance map.} Recurring intervention surfaces are mapped to architectural layers and lifecycle stages, including data and reward design, permission scoping, provenance-aware tool mediation, runtime verification, human escalation, rollback, and continual assurance.
\end{itemize}

\paragraph{Scope, Evidence Base, and Non-Goals}
The analysis is organizational and diagnostic. It does not attempt to verify OpenClaw internals, introduce a new benchmark, or provide a formal safety proof. Its goal is to synthesize existing CUA literature and public deployment patterns through a common analytical frame. The coverage is weighted toward recent CUA work, while retaining earlier studies when they contribute concepts that remain necessary for computer-use reasoning, grounding, execution, or oversight. The selection prioritizes representative systems, benchmarks, surveys, deployment-oriented stacks, and security or privacy studies that materially inform deployed CUA behavior~\cite{osu_wang2024_gui-agents-with-foundation-model,osu_tang2026_clawgui-a-unified-framework-for,osu_kang2026_longhorizonui-a-unified-framewor,osu_zhou2026_webarena-infinity-generating-bro,osu_yuan2026_webforge-breaking-the-realism-re,osu_zhao2026_webpii-benchmarking-visual-pii-d,osu_zhai2026_guide-interpretable-gui-agent-ev,osu_lin2025_cuarewardbench-a-benchmark-for-e}. Adjacent agent-security, tool-use, or governance literature is included only when it clarifies a CUA-relevant mechanism, and is treated as adjacent evidence rather than direct CUA evidence.

\begin{table*}[!t]
\centering
\scriptsize
\caption{Heuristic positioning map for the proposed architecture--lifecycle view within recent CUA literature. The table foregrounds the analytical gap addressed here rather than serving as an exhaustive related-work inventory.}
\label{tab:related_work_comparison}
\renewcommand{\arraystretch}{1.12}
\rowcolors{3}{black!3}{white}
\begin{tabularx}{\textwidth}{>{\raggedright\arraybackslash\bfseries}p{2.7cm} L{3.8cm} L{4.1cm} Y}
\toprule
\rowcolor{black!10}
\textbf{Literature Family} & \textbf{Representative Anchors} & \textbf{What It Best Clarifies} & \textbf{What Remains Less Explicit in Live Deployment} \\
\midrule
Overviews and taxonomies &
GUI and OS-agent overviews~\cite{nguyen2025gui,hu2025agents,tang2025survey,shi2025trustworthygui} &
Method families, platform coverage, benchmark inventories, and high-level ecosystem structure &
How capability formation, authority exposure, failure origin, and control placement interact once the agent leaves the benchmark loop \\
\midrule
Focused survey slices &
RL-enhanced GUI agents, phone automation, WebAgents, and CUA safety/security surveys~\cite{osu_li2025_a-survey-on-gui-agents-with-foun,osu_liu2025_llm-powered-gui-agents-in-phone,osu_ning2025_a-survey-of-webagents-towards-ne,osu_chen2025_a-survey-on-the-safety-and-secur} &
Deeper summaries of one training regime, platform, application surface, or threat family &
How those slices fit into one shared account of deployed reliability across architecture, authority, and lifecycle \\
\midrule
System-building and scaling efforts &
OpenCUA, bBoN, ClawGUI, and LongHorizonUI~\cite{wang2025opencua,gonzalez2025unreasonable,osu_tang2026_clawgui-a-unified-framework-for,osu_kang2026_longhorizonui-a-unified-framewor} &
How CUA capability is being extended through open foundations, scaling, orchestration, and end-to-end stacks &
How similar capability can map to different reliability and risk profiles once tools, permissions, sessions, and oversight are bound in deployment \\
\midrule
Benchmarks, datasets, and environment builders &
Mind2Web, OSWorld, WebForge, and Gym-Anything~\cite{deng2023mind2web,xie2024osworld,osu_yuan2026_webforge-breaking-the-realism-re,osu_aggarwal2026_gym-anything-turn-any-software-i} &
Capability evaluation, realism expansion, synthetic world generation, and broader experimentation infrastructure &
Which failures may be introduced in training, made visible by deployment binding, amplified at runtime, or reopened by post-release drift \\
\midrule
Safety, misuse, and privacy evaluation &
OS-Harm, WAInjectBench, Risky-Bench, and WebPII~\cite{kuntz2025harm,liu2025wainjectbench,zheng2026risky,osu_zhao2026_webpii-benchmarking-visual-pii-d} &
Action-level harm, injection, deployment-grounded risk, and privacy exposure as evaluated in realistic computer-use settings &
How those signals can be mapped to architectural layer, lifecycle stage, and early intervention surfaces rather than remaining benchmark-local observations \\
\midrule
Broad agent security context &
Aegis and broader agent-security overviews~\cite{adapala2025aegis,lazer2026survey} &
Cross-agent governance concepts, protocol design, and security framing beyond one benchmark family &
CUA-specific coupling among GUI observation, execution authority, mixed-trust inputs, human oversight, and continual assurance \\
\midrule
\textbf{Architecture--lifecycle view} &
\textbf{General-purpose CUAs across web, desktop, mobile, and cross-application settings} &
\textbf{An integrated architecture--lifecycle account of capability, risk, privacy, and control} &
\textbf{Clarifies failure origin versus manifestation, maps risk patterns to intervention points, and outlines deployable control surfaces for live-use CUA settings} \\
\bottomrule
\end{tabularx}
\end{table*}

Table~\ref{tab:related_work_comparison} situates the proposed architecture--lifecycle view within the literature, not to rank prior work. It is intentionally selective: the goal is to indicate the analytical scope here rather than to compress the full CUA literature into a single table. Many prior studies already address important aspects of safety, lifecycle, and governance. The contribution here is not to replace those lines of work, but to relate them through a common analytical frame centered on capability formation, authority exposure, failure emergence, and control placement. The remainder develops that frame in sequence. Section~\ref{sec:background} defines the problem space and its main design axes; Section~\ref{sec:framework} introduces the analytical framework; Sections~\ref{sec:architecture} and~\ref{sec:lifecycle} develop the architectural and lifecycle dimensions, respectively; Section~\ref{sec:security} analyzes security and privacy through that joint perspective; and Section~\ref{sec:discussion} draws out the resulting governance and control implications.

\section{Problem Definition and Design Axes}
\label{sec:background}

Before discussing architecture or lifecycle, this section fixes the control problem that a computer-use agent (CUA) is meant to solve. Given a user goal $g$, a constraint set $c$, and a stream of partial observations from a live software environment $\mathcal{E}$, the agent must maintain task-relevant state and emit actions through an authority-bearing interface so that the task progresses without violating user intent or system permissions. The object of study is therefore not interface understanding in isolation. It is control in partially observable, dynamically changing, permission-constrained software environments. The section formalizes that setting and isolates the two design axes that matter most for the following analysis: how the agent constructs executable state from software observations, and how it binds intent to authority-bearing action.

\subsection{Problem Formulation: Software Interaction as Partially Observable Control}
A CUA can be written abstractly as a policy $\pi(a_t \mid h_t, o_t, g, c)$ operating over a software-facing environment $\mathcal{E}$. At step $t$, the agent receives a partial observation $o_t$, updates an internal state $h_t$, and emits an action $a_t \in \mathcal{A}$ through the environment's available control surface. The output is therefore not text alone, but an action with real operational effect. The objective is to reach a task-complete state while preserving user-specified constraints, respecting authority boundaries, and remaining recoverable when uncertainty is high.

What makes this setting distinctive is that software control is only partially observable and only weakly synchronized. A realistic task can move across browsers, operating-system windows, filesystems, terminals, APIs, and tool backends~\cite{wu2024copilot,song2025coact,wang2025opencua,osu_xu2026_mobile-agent-v3-5-multi-platform}. Even with screenshots, DOM or accessibility trees, OCR output, memory traces, and tool responses, relevant state may remain hidden, delayed, or already stale: windows change asynchronously, permission prompts appear after a plan is formed, and tool state may become visible only after an invocation succeeds~\cite{zhou2023webarena,he2024webvoyager,xie2024osworld}.

Three properties follow. First, software environments are semantically dense: a small icon, checkbox, or command argument can carry disproportionate consequences. Second, they are mixed-trust by default: the same observation stream may contain user intent, benign interface content, deceptive prompts, retrieved text, and attacker-controlled or compromised tool output~\cite{liu2025wainjectbench,foerster2026camels,zheng2026risky,yang2025riosworld}. Third, they are layered with authority: the action the model selects is not identical to the permission the system ultimately exercises. This is why CUA behavior is treated here as controlled interaction under partial observability and constrained authority, not as next-step prediction alone.

\subsection{Observation Modalities and Task-Relevant State Representations}
The first major design choice is how raw software observations become task-relevant state. In some environments, DOM trees, accessibility metadata, or view hierarchies provide precise handles for localization and control~\cite{zhou2023webarena}. In others, screenshot-grounded interaction dominates, which turns computer use into a multimodal grounding problem shaped by OCR quality, scale, localization, theme variation, and small-target robustness. Hybrid approaches sit between those extremes by invoking parsers or specialist grounders only when they improve downstream control~\cite{you2024ferret,singh2025trishul,fan2025gui,yuan2025segui}.

Task-relevant state is also broader than whatever is currently visible. A deployed CUA may need to reason over memory traces, retrieved documents, execution logs, tool outputs, file context, and permission status alongside the active interface~\cite{wu2024copilot,song2025coact,luo2025vimo,osu_xie2026_secagent-efficient-mobile-gui-ag}. Environment class then determines which observation assumptions remain viable: web settings stress structure and untrusted content, desktop and OS settings mix GUI state with account authority, mobile settings compress semantics into small targets and permission prompts, and cross-app workflows create handoff errors across surfaces. Benchmarks such as Mind2Web, WebArena, and OSWorld are useful here mainly because they reveal which observation assumptions a design can actually survive~\cite{deng2023mind2web,zhou2023webarena,xie2024osworld}.

The main takeaway of this axis is direct: observation design determines what the agent can know, what it can verify, and what it can trust strongly enough to act on. That is why perception later appears as the first architectural layer rather than as a preprocessing detail.

\subsection{Execution Interfaces, Action Spaces, and Authority Binding}
The second major design choice is how model intent becomes operational authority. Action design is therefore not only a question of granularity. Two action spaces may look equally expressive while exposing very different blast radii, auditability, and rollback properties.

Low-level mouse, keyboard, and touch actions are the most general. They allow the agent to operate almost any visible interface, but they make success depend on precise grounding and stable timing: a coordinate error can become a misclick, and a short delay can become a time-of-check/time-of-use failure. Higher-level actions such as browser APIs, application macros, shell commands, or code execution compress many low-level steps into fewer semantic operations. They often improve efficiency and logging, but they also increase the consequence of each mistake because one invocation may encode a wider authority surface~\cite{wu2024copilot,song2025coact,osu_zhou2025_mai-ui-technical-report-real-wor,osu_yan2025_step-gui-technical-report}.

This is why \emph{action abstraction} and \emph{authority binding} are treated as related but distinct axes. A click is low-level and weakly semantic, but it can still be high-risk if it confirms a destructive operation. A shell command or API invocation is high-level and often more auditable, but it usually carries more direct authority over files, networks, or processes. Modern CUAs increasingly combine both forms of action, which makes execution design inseparable from permission scope, auditability, and recoverability.

The main takeaway of this second axis is equally direct: execution design determines what the agent can cause, how tightly those consequences are bound to authority, and how easy they are to inspect or reverse. That is why execution later appears as a distinct architectural layer rather than as the final step of a generic control loop.

\paragraph{Takeaway}
The central problem of computer use is therefore not only recognizing what is on the screen. It is deciding what should count as executable state, and binding that state to action interfaces that carry real authority. Observation design determines what the agent can know; execution design determines what the agent can cause. Their coupling determines which failures appear first as ambiguity, which appear first as overreach, and why the same error can have very different consequences across architectures and deployment stages. The next section turns that problem definition into an explicit analytical framework.

\section{Analytical Framework}
\label{sec:framework}

The previous section defined the problem space and its main design axes. This section defines how the following analysis reasons about that space. The proposed joint framework combines an \textbf{architectural view} with a \textbf{lifecycle view}. The architectural view explains how a CUA is built across \emph{Perception}, \emph{Decision}, and \emph{Execution}. The lifecycle view explains when capability is created, bound to authority, stressed in live use, and revised over time through \emph{Creation}, \emph{Deployment}, \emph{Operation}, and \emph{Maintenance}. The framework is used as a heuristic analytical lens for organizing literature and locating intervention surfaces in deployed settings~\cite{tang2025survey,osu_li2025_a-survey-on-gui-agents-with-foun,osu_gao2024_generalist-virtual-agents-a-surv}.

It should not be read as a formal causal model, nor as a claim that real systems or incidents fall neatly into mutually exclusive classes. Its purpose is narrower and more practical: to provide a working coordinate system for reasoning about how capability, authority, and oversight become coupled in deployed computer-use settings.

\subsection{Why a Component View Is Not Enough}
Architecture alone remains necessary, but it is incomplete. It tells us how an agent reconstructs interface state, forms plans, and acts. That is enough to compare systems at the component level. It is not enough to explain why the same model behaves well in one environment and fails in another, why one action interface is tolerable under one deployment and materially riskier under another, or why a failure first seen at runtime may in fact originate in data curation, permission binding, or post-release drift.

The lifecycle view adds the missing temporal dimension. \emph{Creation} asks what priors the system learns before release. \emph{Deployment} asks how those priors are coupled to live channels, tools, sessions, and permissions. \emph{Operation} asks how the system behaves under long horizons, mixed-trust inputs, and asynchronous change. \emph{Maintenance} asks whether the same architecture remains valid as interfaces, tools, workflows, and threat models evolve. This second dimension helps separate \emph{where a failure becomes visible} from \emph{where the enabling condition was introduced}. That distinction is a main analytical commitment of the framework.

\begin{table*}[!t]
\centering
\scriptsize
\caption{Architecture--lifecycle coordinate system for the analytical framework. The table maps lifecycle stages to the objects they shape and to the corresponding pressure points in the perception, decision, and execution layers.}
\label{tab:joint_framework}
\renewcommand{\arraystretch}{1.18}
\setlength{\tabcolsep}{3.5pt}
\rowcolors{3}{black!3}{white}
\begin{tabularx}{\textwidth}{>{\raggedright\arraybackslash\bfseries}p{1.35cm} L{1.95cm} L{1.95cm} L{1.95cm} L{1.95cm} Y}
\toprule
\rowcolor{black!10}
\textbf{Stage} & \textbf{Primary Object Shaped} & \multicolumn{3}{c}{\textbf{Where the Tri-Layer Architecture Is Pressured}} & \textbf{Why This Stage Is Analytically Distinct} \\
\rowcolor{black!10}
 &  & \textbf{Perception} & \textbf{Decision} & \textbf{Execution} &  \\
\cmidrule(lr){3-5}
\midrule
Creation & Learned priors, grounding habits, and action vocabulary & What counts as reliable executable state is set by data and supervision & Goal preferences, decomposition habits, and caution thresholds are normalized & Action abstractions and default affordances are chosen before release & Upstream choices can stay latent until real authority is later attached \\
Deployment & Live bindings among observations, tools, sessions, and permissions & Observation channels and provenance are coupled to real software surfaces & Memory, orchestration, and policy mediation enter the live loop & Tools and permissions turn plans into operational authority & The same policy acquires a different risk profile once capability is bound to authority \\
Operation & Active trajectories under uncertainty and mixed trust & Ambiguity, deception, stale state, and missing state appear online & Long-horizon coherence, constraint retention, and verification are stressed & Latency, TOCTOU, and delayed effects shape real consequence & Many failures become visible here even when the enabling condition entered earlier \\
Maintenance & Validity of the model--environment--ecosystem stack & UI drift, localization change, and rendering shift erode grounding fit & Recovery logic, evaluators, and policies can regress after updates & Extensions, permissions, and containment layers drift with the ecosystem & Assurance must survive change, not only one release \\
\bottomrule
\end{tabularx}
\end{table*}

\subsection{What the Joint Framework Claims}
The joint framework makes three claims that organize the rest of the analysis.

\textbf{First, capability is lifecycle-dependent rather than purely model-dependent.} What practitioners often call ``agent ability'' is jointly shaped by representation choices, post-training, execution interfaces, deployment mediation, memory, and recovery mechanisms. Runtime performance is therefore not an intrinsic property of the base model alone.

\textbf{Second, failure origin and failure manifestation are different analytical questions.} A brittle recovery path visible during operation may reflect weak grounding supervision in creation, over-broad or weakly mediated permission scope in deployment, or untested drift in maintenance. Treating every higher-risk action as a runtime reasoning defect hides the real control surface.

\textbf{Third, control placement must follow both layer and stage.} Perception errors, decision errors, and execution errors do not respond to the same controls, and creation-stage controls differ fundamentally from deploy-time, runtime, or maintenance-time ones. Security and governance are therefore not appended after the agent is built; they are about where constraints attach relative to the system's structure and evolution.

Table~\ref{tab:joint_framework} summarizes the working coordinate system used throughout the article. Later tables are narrower instantiations of the same logic rather than competing taxonomies. Whenever later sections compare systems, failures, or controls, the guiding questions remain the same: what stage has changed the object of analysis, and which architectural layer now carries the main pressure?

\paragraph{Assignment Rule}
An issue is assigned primarily by the earliest stage at which it is materially introduced or could still have been constrained. Creation applies when pre-release data, objectives, or action abstractions are the main source of the problem. Deployment applies when live binding decisions about tools, permissions, sessions, observation channels, or mediation dominate. Operation applies when online trajectory evolution under uncertainty is the main source of failure. Maintenance applies when post-release change to the model, environment, or ecosystem is the primary driver. Persistent memory, for example, is a deployment choice; misuse of that memory during a live trajectory is operational; and memory-related regressions after updates belong to maintenance.

\subsection{Deployment Archetypes and an Open-Deployment Example}
To keep the framework concrete without collapsing the analysis into a single case study, three recurring deployment archetypes are useful.

\textbf{Benchmark-centered research agents} are optimized for bounded evaluation. They typically expose narrow authority, controlled observability, and reproducible task structure~\cite{osu_yuan2026_webforge-breaking-the-realism-re,osu_aggarwal2026_gym-anything-turn-any-software-i,osu_zhang2026_clawbench-can-ai-agents-complete}.

\textbf{Consumer or assistant-style deployed agents} sit inside user workflows such as search, browsing, productivity, or messaging. Their defining pressure is that user intent, real accounts, and high-impact actions coexist in one interface loop~\cite{osu_xu2026_mobile-agent-v3-5-multi-platform,osu_zhou2025_mai-ui-technical-report-real-wor,osu_yan2025_step-gui-technical-report}.

\textbf{Self-hosted gateway or tool-binding agents} can make CUA capability available through persistent ingress channels, longer-lived memory, tool registries, or automation layers. Their defining pressure is the coupling among authority, provenance, and ecosystem evolution rather than raw capability alone~\cite{osu_cheng2025_webatlas-an-llm-agent-with-exper,osu_zhang2025_litewebagent-the-open-source-sui,osu_gur2023_a-real-world-webagent-with-plann}.

OpenClaw is used only as a motivating deployment setting. Public materials describe it as a local-first assistant with persistent ingress, tool connectivity, and visible participation in a broader agent community~\cite{openclaw2026website,ibm2026openclaw,chen2026openclaw}. That public framing is treated as an illustrative deployment pattern rather than as verified system evidence: once one deployment surface is presented as combining channels, tools, sessions, and longer-lived context, deployment choices can account for a large share of the resulting risk profile. The analysis remains about the broader class of systems that share a similar structure.

\subsection{How To Read the Rest of the Paper}
The remaining sections form a guided traversal of Table~\ref{tab:joint_framework}. Section~\ref{sec:architecture} develops the architectural dimension by asking how perception, decision, and execution interact. Section~\ref{sec:lifecycle} develops the temporal dimension by asking how capability, failure, and control move across Creation, Deployment, Operation, and Maintenance. Section~\ref{sec:security} maps threats back onto both dimensions. Section~\ref{sec:discussion} then converts that analysis into control surfaces and governance implications. The intended reading is cumulative rather than parallel: each later section reuses the same framework to answer a different layer of the same reliability problem.

\section{Architectural Evolution of Computer-Use Agents: A Tri-Layer Framework}
\label{sec:architecture}

The architectural question in CUAs is simple to state and difficult to answer well: how does an agent convert a changing software environment into justified, executable action? A useful synthesis is structural. A deployed CUA must reconstruct actionable state, preserve task-conditioned intent, and finally translate that intent into authority-bearing operations. This section therefore analyzes CUA architectures through three coupled layers: \emph{Perception}, \emph{Decision}, and \emph{Execution}. The value of this decomposition is not that every system literally contains three modules. It is that the decomposition helps indicate where capability is created, where uncertainty accumulates, and where small errors become operationally expensive~\cite{tang2025survey,shi2025trustworthygui,osu_li2025_a-survey-on-gui-agents-with-foun}.

The recent literature makes this tri-layer view increasingly necessary. Perception work spans structured web grounding, parser-augmented visual understanding, native screenshot-grounded policies, and newer grounding methods that adapt zoom, attention, or symbolic structure to interface complexity~\cite{lu2024omniparser,hui2025winclick,singh2025trishul,fan2025gui,osu_pei2026_adazoom-gui-adaptive-zoom-based,yuan2025segui}. Planning and control work now covers memory-augmented reasoning, long-horizon decomposition, rollback, reflection, and hybrid planners that mix GUI control with programmatic execution~\cite{zheng2023synapse,zhang2025ufo,song2025coact,osu_he2026_ee-mcp-self-evolving-mcp-gui-age,osu_jiang2026_treecua-efficiently-scaling-gui,osu_yu2026_graphpilot-gui-task-automation-w}. Execution work is widening at the same time, from atomic GUI actions to code-as-action, dexterous control, and bundled routines, which means architectural choice is now inseparable from authority design~\cite{wu2024copilot,jiang2025appagentx,mu2025gui,osu_hu2025_showui-flow-based-generative-mod,osu_yu2025_polyskill-learning-generalizable}. The field does not lack methods; instead, recent capability gains often shift risk across layers rather than remove it.

\begin{figure*}[t]
\centering
\includegraphics[width=\textwidth]{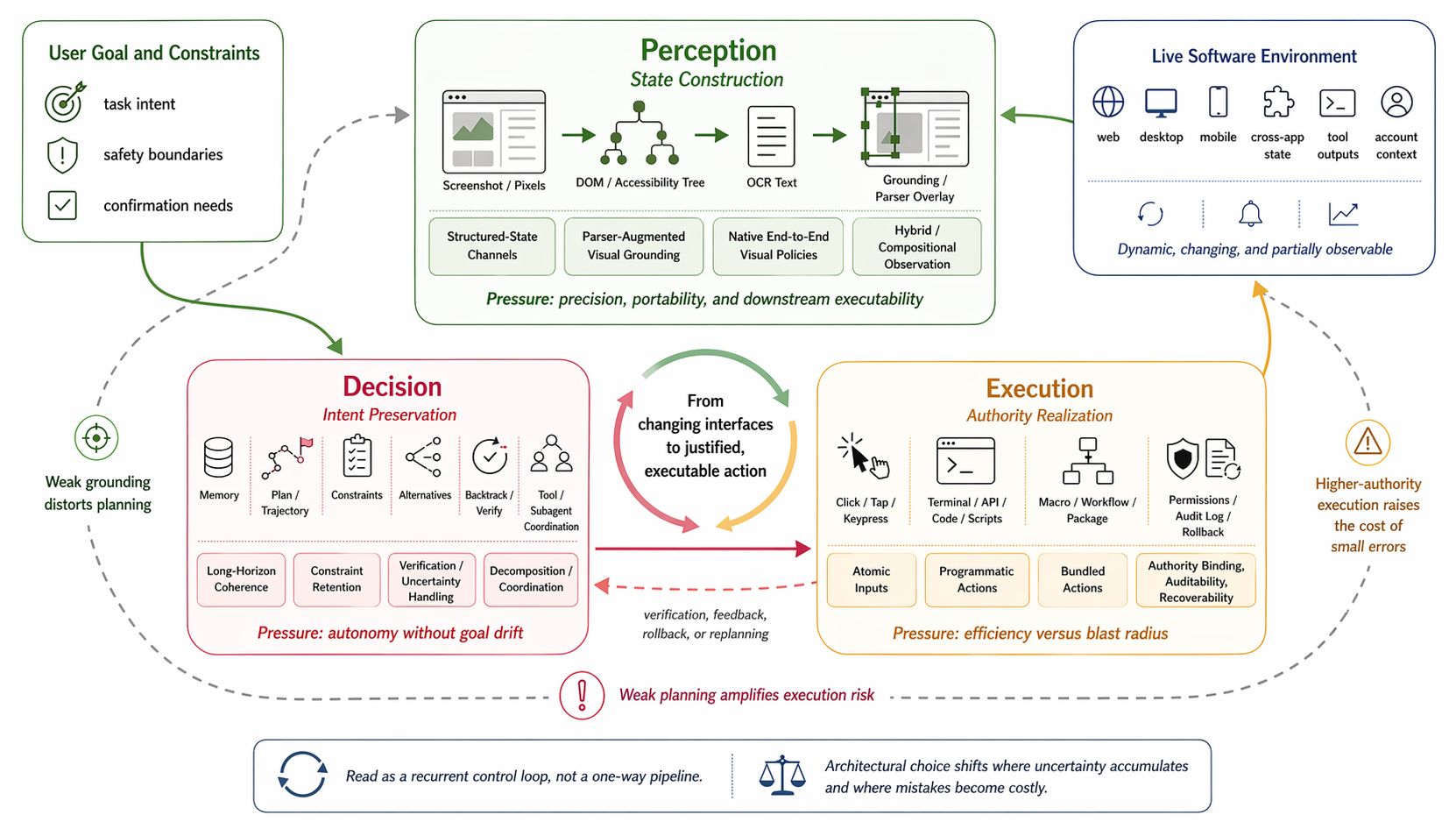}
\caption{Tri-layer pressure map for deployed CUAs. The figure frames deployed CUA behavior as a recurrent control loop between user goals and constraints and a live, partially observable software environment. Perception reconstructs operationally trustworthy state, decision maintains task-conditioned intent under long-horizon pressure, and execution converts the resulting trajectory into authority-bearing action.}
\label{fig:tri_layer_architecture}
\end{figure*}

Figure~\ref{fig:tri_layer_architecture} depicts a recurrent control loop rather than a one-way pipeline. The live environment supplies dynamic and partially observable state to perception, user goals and constraints shape decision, and execution not only acts on the environment but can also return verification, feedback, rollback, or replanning signals to decision. Perception shapes what the system can treat as actionable state. Decision shapes whether that state is converted into a coherent and constrained trajectory. Execution shapes how much authority is exercised when that trajectory is enacted. Failure pressure can propagate accordingly: weak grounding can distort planning, weak planning can amplify execution risk, and high-authority execution can raise the cost of even small perceptual or decisional mistakes.

\begin{table*}[!t]
\centering
\scriptsize
\caption{Architectural archetypes in contemporary CUAs. The table summarizes recurring design patterns, their primary capability gains, the layers they chiefly strengthen, and the reliability pressures they introduce in deployed stacks.}
\label{tab:method_taxonomy}
\renewcommand{\arraystretch}{1.12}
\setlength{\tabcolsep}{3.5pt}
\rowcolors{3}{black!3}{white}
\begin{tabularx}{\textwidth}{>{\raggedright\arraybackslash\bfseries}p{1.9cm} L{2.7cm} L{2.15cm} L{1.75cm} Y}
\toprule
\rowcolor{black!10}
\textbf{Archetype} & \textbf{Representative Systems} & \textbf{Main Capability Gain} & \textbf{Layer Chiefly Strengthened} & \textbf{Main Reliability Pressure} \\
\midrule
Structured-state web agents & Mind2Web and WebArena-style agents~\cite{deng2023mind2web,zhou2023webarena} & Precise element grounding and deterministic references when structure is reliable & Perception & Portability narrows once dependable structure disappears or becomes weakly exposed \\
Parser-augmented visual agents & OmniParser, WebVoyager, and SeeClick~\cite{lu2024omniparser,he2024webvoyager,cheng2024seeclick} & Recover action handles from rendered interfaces without full dependence on raw DOM exposure & Perception & Parser errors can propagate downstream as confident but distorted action state \\
Native end-to-end visual agents & WinClick, Mobile-Agent, and MAI-UI~\cite{hui2025winclick,wang2024mobile,osu_zhou2025_mai-ui-technical-report-real-wor} & Broad cross-platform reach from screenshot-grounded policies & Perception + Decision & Failures become harder to localize, and small-target robustness remains costly \\
Memory-augmented planners & Synapse, WebATLAS, and ColorBrowserAgent~\cite{zheng2023synapse,osu_cheng2025_webatlas-an-llm-agent-with-exper,osu_wang2026_colorbrowseragent-complex-long-h} & Stronger long-horizon coherence and task continuity & Decision & Stale summaries, poisoned memory, or dropped constraints can steer later steps \\
Hybrid orchestration agents & UFO, GraphPilot, and LiteWebAgent~\cite{zhang2025ufo,osu_yu2026_graphpilot-gui-task-automation-w,osu_zhang2025_litewebagent-the-open-source-sui} & Decompose work across applications, tools, or specialized roles & Decision + Execution & Coordination, attribution, and interface-selection errors move burden across components \\
Hybrid execution agents & OS-Copilot, CoAct-1, and AppAgentX~\cite{wu2024copilot,song2025coact,jiang2025appagentx} & Replace brittle GUI sequences with code, tools, or reusable routines & Execution & Stronger authority binding and harder rollback enlarge the consequence of mistakes \\
\bottomrule
\end{tabularx}
\end{table*}

Table~\ref{tab:method_taxonomy} summarizes heuristic architectural bets rather than ranking individual systems. The same deployed CUA may combine several rows at once, which is precisely why capability gains often reappear later as cross-layer reliability pressure.

\subsection{Perception Layer: From Raw Interfaces to Actionable State}

The perception layer answers the first architectural question: what does the agent believe the interface contains, and how trustworthy is that belief for downstream action? In CUAs, perception is not only recognition. It is the construction of operational state from screenshots, DOM trees, accessibility metadata, OCR output, parser-derived layouts, and retrieved contextual traces. The design problem is therefore not simply to maximize visual accuracy. It is to decide what form of state remains precise enough to act on and portable enough to survive across environments.

The literature now falls into four recurring families. \textbf{Structured-state agents} rely on DOM, accessibility, or view-hierarchy signals. Their strength is precision when the environment exposes reliable structure, which is why Mind2Web and WebArena remain important reference points for disciplined web control~\cite{deng2023mind2web,zhou2023webarena}. Their weakness is portability: once interfaces become rendering-heavy, streamed, or weakly instrumented, the same structural advantage can disappear.

\textbf{Parser-augmented visual agents} recover part of that structure from rendered interfaces. Systems such as OmniParser, ScreenAI, SeeClick, Ferret-UI, TRISHUL, and newer complete-screen parsing approaches add OCR, icon captioning, region proposals, or layout parsing before downstream reasoning~\cite{lu2024omniparser,baechler2024screenai,cheng2024seeclick,you2024ferret,singh2025trishul,gurbuz2026moving}. This family is attractive because it retains visual generality while recovering some of the handles that make execution easier. Its main failure mode is familiar: the parser becomes a bottleneck, and downstream reasoning can remain confidently wrong when the parsed state is incomplete or distorted.

\textbf{Native end-to-end visual agents} push farther toward portability by leaving more of grounding inside one multimodal policy. CogAgent, Fuyu-style VLMs, Qwen-VL style systems, WinClick, MAI-UI, Step-GUI, Mobile-Agent-v3.5, and high-resolution-aware agents such as AFRAgent illustrate this direction~\cite{hong2024cogagent,fuyu-8b,bai2023versatile,hui2025winclick,osu_zhou2025_mai-ui-technical-report-real-wor,osu_yan2025_step-gui-technical-report,osu_xu2026_mobile-agent-v3-5-multi-platform,osu_anand2025_afragent-an-adaptive-feature-ren}. The gain is broad environment coverage. The cost is entanglement. OCR, localization, affordance inference, and actionability all live inside one representation, which makes it harder to know whether a failure arose from vision, reasoning, or the coupling between the two.

\textbf{Hybrid or compositional grounding agents} mix several channels rather than committing to one. They use structured state when it is available, screenshots when it is not, and specialist grounders only when those tools improve downstream control~\cite{wu2024copilot,zhang2025ufo,fan2025gui,osu_andreux2025_surfer-2-the-next-generation-of}. This family is especially important for cross-platform and open-deployment settings because the observation channel itself may vary within a single task.

These families are best compared along four axes: element-grounding precision, small-target robustness, robustness to scale/theme/localization changes, and downstream executability. No current approach simultaneously maximizes all four. Benchmarks such as ScreenSpot-Pro help make the trade-off concrete by indicating that high-resolution professional interfaces still punish small-target grounding even when broader GUI capability looks competitive elsewhere~\cite{li2025screenspotpro}. The most precise observation channel is rarely the most portable, and the most portable channel is rarely the easiest to verify. That tension is why perception in CUAs should be judged not by recognition accuracy alone, but by whether the resulting state remains trustworthy enough for safe execution.

\paragraph{Observation-channel mismatch}
One under-discussed architectural risk is mismatch between the observation channel assumed during training and the one encountered in deployment. A model trained on parser-normalized or DOM-rich state can look strong on benchmarks yet fail in screenshot-only or streamed settings. A screenshot-trained model can make the opposite mistake by ignoring structure that would have improved precision. The architectural point is that a policy can look transferable while the real grounding problem has already changed underneath it.

This classification remains heuristic rather than exhaustive. Many deployed systems mix structured state, screenshots, parsers, and specialist grounders within one task, so the perception families above capture dominant observation bets rather than clean system boundaries.

\subsection{Decision Layer: Preserving Intent Under Long-Horizon Pressure}

If perception answers ``what is happening now,'' the decision layer answers ``what should happen next, and how can the system verify that it still matches the user's goal?'' In deployed CUAs, this is where autonomy becomes either stable or brittle. The core difficulty is not merely local next-step selection. It is preserving task-conditioned intent across long trajectories, incomplete observations, changing interfaces, and sometimes hostile runtime content.

Four pressures organize the literature here. The first is \textbf{long-horizon coherence}: can the system keep the task moving in the right direction across many screens and tool calls? The second is \textbf{constraint retention}: can it preserve instructions such as ``draft, do not send'' or ``ask before deleting'' after dozens of intermediate steps? The third is \textbf{verification and uncertainty handling}: does it know when to check, backtrack, or ask for clarification? The fourth is \textbf{decomposition}: does it solve the task in one policy loop, or distribute work across roles, tools, or subagents?

Memory-augmented systems are an early answer to the coherence problem. Synapse illustrated that state abstraction, trajectory exemplars, and retrieval can materially stabilize computer-control behavior under limited context~\cite{zheng2023synapse}. AppAgent and InfiGUIAgent push the same idea toward reusable operating knowledge rather than disposable context~\cite{zhang2025appagent,liu2025infiguiagent}. Newer systems such as WebATLAS, SecAgent, ColorBrowserAgent, and anchored-memory benchmarks such as AndroTMem make a similar point from different directions by treating summarization, simulation, or anchored recall as first-class planning supports rather than optional add-ons~\cite{osu_cheng2025_webatlas-an-llm-agent-with-exper,osu_xie2026_secagent-efficient-mobile-gui-ag,osu_wang2026_colorbrowseragent-complex-long-h,osu_shi2026_androtmem-from-interaction-traje}. OSWorld evaluations are consistent with the same concern: long tasks can degrade quickly without some mechanism for compression, recall, or structured recovery~\cite{xie2024osworld}. These systems matter because they suggest that CUA performance is not only a function of local reasoning quality. It also depends on how continuity is carried forward.

Constraint retention is the more safety-critical version of the same problem. A CUA can appear productive while gradually dropping confirmation requirements, user preferences, or exposure limits. Reflection-oriented work and safe-planning or interruptibility evaluations are therefore informative not only because they improve task completion, but because they ask whether the original task boundary stays attached to later steps~\cite{wu2025guireflection,li2025mobileuse,chen2026lps,osu_zou2026_when-users-change-their-mind-eva,osu_yang2026_guide-a-benchmark-for-understand}. The more concerning failure mode is often not that the model stops progressing. It is that it keeps progressing after silently forgetting what should have constrained it.

Verification and uncertainty handling are what keep that drift from becoming irreversible. General agent methods such as ReAct, Tree of Thoughts, and Reflexion contributed three ideas that recur in CUA planners: interleaving reasoning with action, branching over alternatives, and using prior errors as planning signals~\cite{yao2022react,yao2023tree,shinn2023reflexion}. WebVoyager, You Only Look at Screens, ReInAgent, Mobile-Agent, BacktrackAgent, TreeCUA, stable-planner modules, and action-effect verification systems carry those ideas into live interfaces through backtracking, chain-of-action reasoning, active questioning, replanning, explicit recovery after detected errors, and post-action verification~\cite{he2024webvoyager,zhang2024you,jia2025reinagent,wang2024mobile,osu_wu2025_backtrackagent-enhancing-gui-age,osu_jiang2026_treecua-efficiently-scaling-gui,osu_mo2025_building-a-stable-planner-an-ext,osu_zhang2026_don-t-act-blindly-robust-gui-aut}. The literature does not point to one universally superior planner. It does suggest, however, that robust computer use cannot remain purely feedforward once trajectories become long and mixed-trust.

Decomposition offers a second route to scale. UFO separates application routing from local action selection, while CoAct-1 coordinates GUI operation with programmatic execution through a central planner~\cite{zhang2025ufo,song2025coact}. EE-MCP, GraphPilot, LiteWebAgent, Agent-SAMA, PolySkill, and earlier planning-oriented web agents extend the same design space by balancing GUI interaction against tool calls, reusable skills, finite-state planning, or program synthesis, which makes the coordination problem partly an interface-selection and skill-selection problem rather than only a planning problem~\cite{osu_he2026_ee-mcp-self-evolving-mcp-gui-age,osu_yu2026_graphpilot-gui-task-automation-w,osu_zhang2025_litewebagent-the-open-source-sui,osu_guo2025_agent-sama-state-aware-mobile-as,osu_yu2025_polyskill-learning-generalizable,osu_gur2023_a-real-world-webagent-with-plann}. These designs can make complex workflows tractable, but they do not eliminate difficulty. They move difficulty from single-policy context management into coordination, attribution, and trust management across components.

The architectural lesson is therefore narrow but important: decision quality in CUAs is not the ability to choose the next plausible click. It is the ability to preserve user-intended constraints while converting uncertain, partial state into a trajectory that remains coherent over time. Stronger planning without stronger verification usually means stronger drift.

This decision-layer classification also has limits. Real systems often fuse memory, reflection, skills, planners, and verification into one controller, so the categories above capture recurring planning pressures rather than strict module types.

\subsection{Execution Layer: Action Abstraction, Authority, and Recoverability}

The execution layer is where the reliability argument becomes concrete. A CUA does not merely choose what should happen. It acts through a specific interface that determines how much authority the decision carries, how easy the action is to audit, and how recoverable the outcome remains when something goes wrong. In other words, execution is where intent becomes consequence.

Three execution styles recur across the literature. \textbf{Atomic input execution} uses clicks, drags, taps, and keystrokes. It is maximally general and often the only option when no trusted higher-level interface exists. Cradle and AndroidEnv illustrate the breadth of this style, while many visual desktop and mobile agents inherit the same strengths and weaknesses~\cite{tan2024cradle,toyama2021androidenv,wang2025ui,osu_hu2025_showui-flow-based-generative-mod}. Its advantage is portability. Its failure modes are misclicks, action rebinding, clickjacking, and time-of-check/time-of-use gaps.

\textbf{Programmatic execution} compresses many low-level steps into semantically richer operations. OS-Copilot and CoAct-1 illustrate how file operations, application macros, shell commands, or code execution can bypass brittle GUI sequences in complex workflows~\cite{wu2024copilot,song2025coact}. GUI-360, MAI-UI, and Step-GUI add complementary evidence that hybrid GUI+API or GUI+MCP action spaces are a recurring part of the current CUA design landscape rather than isolated systems choices~\cite{mu2025gui,osu_zhou2025_mai-ui-technical-report-real-wor,osu_yan2025_step-gui-technical-report}. The gain is speed, auditability, and leverage. The trade-off is stronger authority binding: one invocation can modify files, alter configurations, or contact external systems directly.

\textbf{Bundled or macro-action execution} sits between those extremes. AppAgentX is a direct example because it evolves recurrent action sequences into higher-level routines that substitute for repeated low-level interaction~\cite{jiang2025appagentx}. PolySkill provides a related skill-abstraction view in which reusable polymorphic skills stand in for repeated interaction fragments across tasks~\cite{osu_yu2025_polyskill-learning-generalizable}. This can reduce planning burden and improve efficiency, but it also makes failure harder to localize because several consequential substeps may be bundled into one command. More general screenshot-grounded systems such as ScreenAgent and OmegaUse reinforce the field's movement toward broad end-to-end GUI control, but they do not resolve on their own where rollback and verification boundaries should sit~\cite{niu2024screenagent,zhang2026omegause}.

These three styles are best compared through two coupled questions: how abstract is the action, and how much authority is bound to it? That distinction clarifies why shell execution belongs in the execution layer rather than in GUI control. It is a high-authority interface, not a visual interaction primitive. It also clarifies why recoverability deserves its own place in CUA design. Atomic inputs are easy to interrupt but hard to replay semantically. Programmatic actions are easier to log and replay but often harder to roll back safely. Bundled actions reduce step count while blurring rollback boundaries. For deployed CUAs, recoverability can matter more than raw task completion because it determines whether a failure remains local or becomes irreversible.

This execution grouping should not be read as a set of mutually exclusive modes. In practice, many deployed CUAs mix atomic, programmatic, and bundled execution within the same workflow, which is exactly why authority binding and rollback boundaries remain hard to reason about.

\subsection{Cross-Layer Coupling and Architectural Implications}

The three layers should be read together. Perception determines what state is available for planning. Decision determines whether that state is converted into a coherent and constrained trajectory. Execution determines how much real authority that trajectory acquires. Better performance in one layer can therefore shift pressure to another: more portable perception can increase ambiguity, stronger autonomy can increase the need for verification, and more efficient execution can enlarge the blast radius of mistakes.

This is why architecture already contains a latent governance question. A system that combines ambiguous perception, long-horizon autonomy, and high-authority execution is not only an engineering design. It can also create a distinctive configuration of risk. Reliability problems in CUAs often become visible as cross-layer coupling before they appear as isolated bugs~\cite{shi2025trustworthygui,yang2025macosworld}. The tri-layer framework is therefore the structural backbone of the analysis. It explains where capability resides, where ambiguity enters, and where authority is exercised. The next section adds the missing temporal backbone by asking how those same layers are pressured differently across Creation, Deployment, Operation, and Maintenance.

\section{A Lifecycle Framework for Computer-Use Agents: From Capability Formation to Continual Adaptation}
\label{sec:lifecycle}

The previous section explained what a CUA is made of. This section explains how that architecture becomes a live system over time. This distinction matters because the same visible failure can emerge from different upstream causes. A misleading click may reflect weak grounding supervision during training, unsafe permission binding at release, brittle recovery under runtime drift, or a regression introduced during maintenance. The lifecycle view is therefore used not to retell the development timeline, but to distinguish when capability is formed, when authority is attached, when risk first becomes visible, and where effective controls can still enter.

Recent benchmarks and environments make this lifecycle sensitivity easier to see. Newer evaluations increasingly include personalized mobile interaction, privacy-sensitive workflows, collaborative assistance, verification-centered replay, software environments generated at scale, and human-like long-horizon behavior~\cite{osu_ramesh2026_websp-eval-evaluating-web-agents,osu_zhou2026_webarena-infinity-generating-bro,osu_nie2026_pspa-bench-a-personalized-benchm,osu_yuan2026_webforge-breaking-the-realism-re,osu_zhu2026_turing-test-on-screen-a-benchmar,osu_yang2026_guide-a-benchmark-for-understand,osu_huq2026_modeling-distinct-human-interact}. These settings do not by themselves establish a lifecycle theory. They do suggest, however, that capability formation, deployment binding, runtime pressure, and maintenance drift leave different empirical signatures once agents are evaluated outside narrow one-shot tasks.

\begin{figure*}[t]
\centering
\includegraphics[width=\textwidth]{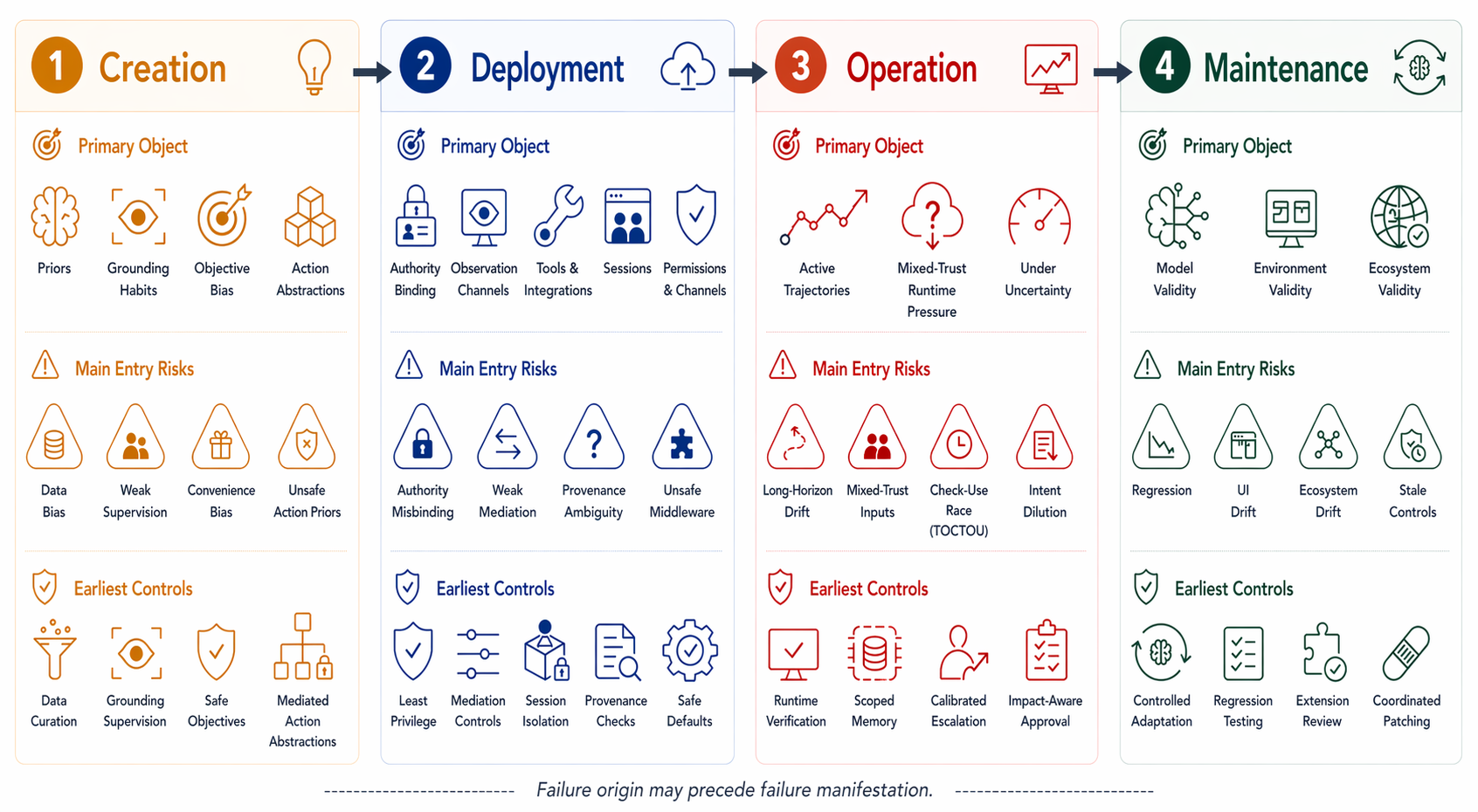}
\caption{Revised four-stage lifecycle framework for deployed CUAs. Each stage is organized into three aligned bands: the \emph{primary object} shaped at that stage, the \emph{main entry risks} that can first be introduced there, and the \emph{earliest controls} that may still change downstream outcomes. The bottom guide emphasizes the main analytical point: failures may be introduced earlier and become visible later, so failure origin can precede failure manifestation.}
\label{fig:lifecycle_taxonomy}
\end{figure*}

Throughout this section, each stage is read through the same analytical template: \emph{What primary object does this stage shape? What risks can first enter here? Why might they only become visible later? Where is the earliest control surface that can still matter?} Figure~\ref{fig:lifecycle_taxonomy} makes that template explicit by organizing every stage into three aligned bands: \emph{Primary Object}, \emph{Main Entry Risks}, and \emph{Earliest Controls}. The categories inside each band are not meant as exhaustive engineering checklists. They summarize the main ways in which a stage can change the eventual behavior of a deployed CUA.

\subsection{Why These Four Stages and Why These Categories}

The four-stage split is intentionally minimal. \emph{Creation} is separated out because it shapes the policy's priors before the system meets any live account or interface. \emph{Deployment} is distinct because it changes the binding between learned capability and real authority. \emph{Operation} is where the active trajectory is exposed to temporal pressure, mixed-trust inputs, and partial observability. \emph{Maintenance} is separate because post-release drift affects the model, the environment, and the surrounding ecosystem differently. In the revised figure, those differences are captured as different \emph{primary objects}: priors and grounding habits in creation, authority-bearing bindings in deployment, active trajectories under uncertainty in operation, and validity of the model--environment--ecosystem stack in maintenance.

This separation is analytically useful because coarser decompositions hide important differences in failure origin. If creation and deployment are merged, training-time bias becomes hard to distinguish from over-authority introduced only at release. If operation and maintenance are merged, live runtime stress is confused with regressions or ecosystem drift. The four stages are therefore analytical categories tied to when different enabling conditions are introduced, not just milestones on a project timeline. The revised diagram also makes the analytical sequence more concrete: once the stage-specific object is identified, the next question is what risks first enter there, and then which controls can still act early enough to matter.

The subcategories inside each stage follow the same logic. \emph{Creation} is decomposed by the mechanisms that shape priors before release, such as data bias, weak supervision, objective bias, and unsafe action abstractions. \emph{Deployment} is decomposed by the bindings through which capability acquires operational effect, including authority binding, observation channels, tools, sessions, and permissions. \emph{Operation} is decomposed by the runtime pressures that can turn local uncertainty into higher-risk trajectories, such as long-horizon drift, mixed-trust inputs, TOCTOU, and intent dilution. \emph{Maintenance} is decomposed by the post-release objects that drift independently, including the model, the environment, and the surrounding ecosystem. Read this way, the lifecycle framework is not an expanded project timeline. It is a map from failure origin to intervention point.

The stage boundaries are operational rather than merely conceptual. Creation is used when a problem is primarily driven by pre-release data, objective design, or action abstraction. Deployment is used when live binding decisions about tools, permissions, sessions, observation channels, or mediation dominate. Operation is used when online trajectory evolution under uncertainty is the main source of failure. Maintenance is used when post-release change to the model, environment, or ecosystem becomes the primary driver. Persistent memory, for example, is a deployment choice; misuse of that memory during a live trajectory is an operational phenomenon; and memory-related regressions after updates belong to maintenance.

\subsection{Creation: Building Priors Before Release}

Creation shapes the object that later gets deployed. It determines what counts as actionable state, what trajectories look normal, and what trade-offs the agent learns when task success competes with caution, confirmation, or recoverability. Failures that enter here may remain latent for a long time. They often surface only after the system is trusted with real authority. That is precisely why this stage matters.

\paragraph{Training data source and trajectory quality}
Training data determines which behaviors and edge cases are even visible to the learner. Human trajectory corpora such as AITW, web interaction datasets such as Mind2Web, and generalized agent tuning pipelines suggested that CUAs benefit from task-conditioned action traces rather than generic multimodal pairs~\cite{rawles2023androidinthewild,deng2023mind2web,zeng2024agenttuning}. Newer resources such as OpenCUA, TongUI, WebChain, MolmoWeb, and OS-Genesis increase scale, platform diversity, interface coverage, or synthetic trajectory breadth~\cite{wang2025opencua,zhang2025tongui,osu_fan2026_webchain-a-large-scale-human-ann,osu_gupta2026_molmoweb-open-visual-web-agent-a,osu_sun2024_os-genesis-automating-gui-agent}. Yet scale alone does not solve the main bias problem. Human traces often encode hesitation, confirmation, and recovery. Synthetic or tutorial-derived traces scale faster, but they can overrepresent clean completion paths and underrepresent ambiguity or conservative stopping. Later runtime failures such as overconfident continuation or brittle recovery are often seeded here before they are ever observed live.

\paragraph{Grounding supervision and observation priors}
Creation also decides what the model is taught to trust as executable state. Structured supervision can improve precise element grounding, whereas screenshot-first training improves portability. Systems such as CogAgent, ScreenAI, SeeClick, WinClick, Ferret-UI, TRISHUL, and newer screen-parsing or exploration-based grounding pipelines suggest that downstream reliability can still depend heavily on that choice~\cite{hong2024cogagent,baechler2024screenai,cheng2024seeclick,hui2025winclick,you2024ferret,singh2025trishul,gurbuz2026moving,fan2025gui,yuan2025segui}. Explicit grounding supervision makes the operational target clearer. Weak supervision can still produce good benchmark performance, but it often leaves the system relying on unstable salience, parser artifacts, or theme-specific cues that later fail in deployment.

\paragraph{Objective design, reward shaping, and action priors}
Objective design determines what the model is implicitly rewarded to optimize. Work on UI-R1, GUI-R1, ProgRM, Web-Shepherd, UI-Genie, MagicGUI-RMS, Video-Based Reward Modeling, and CUARewardBench suggests that reward-shaped post-training and reward-model design can materially shift action efficiency, verification quality, and task completion behavior~\cite{lu2025ui,luo2025gui,osu_zhang2025_progrm-build-better-gui-agents-w,osu_chae2025_web-shepherd-advancing-prms-for,osu_xiao2025_ui-genie-a-self-improving-approa,osu_li2026_magicgui-rms-a-multi-agent-rewar,osu_song2026_video-based-reward-modeling-for,osu_lin2025_cuarewardbench-a-benchmark-for-e}. The lifecycle implication is an inference rather than a direct claim of those papers: if training increasingly rewards low action count, fast completion, or shortcut-taking without parallel incentives for confirmation and recoverability, then a convenience bias may be introduced before any live account is touched. Action abstractions matter for the same reason. If the action vocabulary normalizes broad, weakly mediated authority, that authority may already be treated as ordinary at creation time.

\paragraph{Interactive environments as pre-release stress}
Creation is also where the field can expose delayed consequences before release. Benchmarks such as WebArena, WebForge, Gym-Anything, ClawBench, WorkArena++, and OSWorld matter not only because they measure capability, but because they reveal multi-step recovery, delayed effects, open-environment noise, and broader software coverage while the system is still under development~\cite{zhou2023webarena,osu_yuan2026_webforge-breaking-the-realism-re,osu_aggarwal2026_gym-anything-turn-any-software-i,osu_zhang2026_clawbench-can-ai-agents-complete,osu_boisvert2024_workarena-towards-compositional,xie2024osworld}. They are best read as pre-release stress instruments that surface creation-stage weakness earlier than closed instruction-following benchmarks can.

\paragraph{Why creation-stage failures stay hidden}
The reason creation matters so much is that its failures often look fine until authority is attached. A model can appear capable in benchmark loops while already carrying the wrong observation priors, the wrong convenience bias, or the wrong action habits. The earliest effective control point for creation-stage failures therefore lies in data design, supervision quality, objective shaping, and safe action abstraction.

\subsection{Deployment: Binding Capability to Authority}
\label{sec:deployment}

Deployment changes the object of analysis from a capable policy into a consequential system. The model is now connected to real observation channels, real tools, real sessions, real permissions, and real users. Capability is not merely exposed to the world here; it is bound to authority. This is why deployment deserves its own stage rather than being collapsed into generic release engineering.

\paragraph{Observation binding}
Observation binding determines what state reaches the model once it leaves the training environment. Screenshot-only deployments maximize reach but accept ambiguity, latency, and mixed-trust content. Screenshot-first systems such as Cradle, Mobile-Agent, Surfer 2, and Mobile-Agent-v3.5 illustrate that reach~\cite{tan2024cradle,wang2024mobile,osu_andreux2025_surfer-2-the-next-generation-of,osu_xu2026_mobile-agent-v3-5-multi-platform}. Controlled environments such as WebArena reduce ambiguity by exposing cleaner state or more deterministic execution paths~\cite{zhou2023webarena}. The key trade-off is not simply convenience. It is portability versus controllability.

\paragraph{Tool binding}
Tool binding determines how many external interfaces the agent can call and through what mediation. Systems such as OS-Copilot, UFO, CoAct-1, GraphPilot, LiteWebAgent, MAI-UI, and Step-GUI illustrate how tool augmentation can expand what CUAs can achieve, especially across applications and action channels~\cite{wu2024copilot,zhang2025ufo,song2025coact,osu_yu2026_graphpilot-gui-task-automation-w,osu_zhang2025_litewebagent-the-open-source-sui,osu_zhou2025_mai-ui-technical-report-real-wor,osu_yan2025_step-gui-technical-report}. At the same time, every new tool call introduces another trust boundary. Tool outputs influence reasoning, tool invocations affect authority, and the tool ecosystem itself becomes part of the deployment surface.

\paragraph{Permission binding}
Permission scope is where abstract capability turns into blast radius. A model that can view or modify local files, send messages, change settings, or install software is not merely a stronger benchmark policy. It can function as a live principal acting under inherited authority. Least privilege, sandboxing, and execution mediation therefore belong to deployment by design, not only to incident response.

\paragraph{Memory, session, and channel binding}
Deployment also decides whether context is ephemeral, task-local, cross-session, or persistent across users and channels. Persistent memory can improve continuity and make assistants feel more competent. It can also enlarge the security and privacy surface because old state remains operationally available long after the original task boundary has passed. Personalized and socially embedded benchmarks such as PSPA-Bench, KnowU-Bench, and CowCorpus help make that shift visible by foregrounding long-lived user context, collaborative intervention, and context carry-over across tasks~\cite{osu_nie2026_pspa-bench-a-personalized-benchm,osu_chen2026_knowu-bench-towards-interactive,osu_huq2026_modeling-distinct-human-interact}. The same is true of channel exposure. Messaging routes, automation triggers, browser sessions, shared desktops, and internal tools differ sharply in provenance and trust assumptions.

\paragraph{Why deployment changes the risk profile}
Public materials describe OpenClaw as a gateway-style assistant with persistent ingress, local-first execution, tool connectivity, and longer-lived context~\cite{openclaw2026website,ibm2026openclaw}. That motivating deployment setting is used only as an illustrative deployment pattern rather than as verified system evidence: once one deployment surface is presented as binding channels, tools, sessions, and permissions together, deployment configuration can account for a large share of the resulting risk profile. Comparable design pressures can also be seen in MCP-enabled assistants, local automation gateways, and other multi-channel CUA deployments~\cite{zhang2025mcp,li2025dissonances,wu2024copilot}. The earliest effective control point for deployment-stage failures therefore lies in mediation, permission scoping, session isolation, provenance, and safe defaults.

\subsection{Operation: Runtime Pressure, Mixed Trust, and Oversight}
\label{sec:operation}

Operation is where the deployed system is stressed under live time. The architecture has been chosen, authority has been bound, and the agent now has to remain coherent while the environment changes, the task unfolds, and new content enters the loop. Many failures become visible here even when they did not originate here. That is why runtime incidents should not automatically be read as runtime causes.

\paragraph{Long-horizon drift and stale context}
One source of operational failure is trajectory degradation. WorldGUI, MobileUse, ActionEngine, the Amazing Agent Race, ClawBench, MobileWorldBench, and AndroTMem suggest in different ways that step-wise competence does not automatically produce reliable long-horizon behavior~\cite{zhao2025worldgui,li2025mobileuse,zhong2026actionengine,osu_kim2026_the-amazing-agent-race-strong-to,osu_zhang2026_clawbench-can-ai-agents-complete,osu_li2025_mobileworldbench-towards-semanti,osu_shi2026_androtmem-from-interaction-traje}. An agent can keep taking plausible local steps while losing the original goal, working off stale state, or repeating low-value recovery loops. This is often the main path by which capable-looking systems become operationally unreliable.

\paragraph{Mixed-trust runtime inputs}
Another source is mixed-trust input. During live use, the agent consumes page text, retrieved snippets, messages, document content, and tool outputs that do not share one trust level. Runtime operation is therefore where injection and deceptive content stop being abstract evaluation categories and become active task conditions~\cite{osu_wang2025_webinject-prompt-injection-attac,osu_yang2025_in-context-defense-in-computer-a,osu_chen2025_evaluating-the-robustness-of-mul}.

\paragraph{TOCTOU and delayed execution}
Operation also exposes the temporal gap between decision and effect. Interface state can change after the plan is formed but before the click lands, the macro runs, or the command executes. When this happens, a locally sensible action can become globally misbound or otherwise unsafe without the goal itself ever changing. This is one reason execution verification belongs in operation rather than only in design.

\paragraph{User-intent dilution in persistent workflows}
Once sessions persist across tasks, channels, or users, the system can blur ownership and scope. Personalized long-lived evaluations and deployment-oriented analyses are consistent with that concern in settings where memory, task identity, and ingress are modeled as longer-lived than a single interaction~\cite{wang2026assistant,wang2026your}. The issue is not only security. It is whether user intent remains the dominant organizing constraint in a long-lived runtime context.

\paragraph{Why runtime control must be explicit}
These pressures are why operation needs an explicit oversight ladder rather than vague references to ``human in the loop.'' In increasing order of control strength, that ladder typically includes logging only, post-action verification, plan preview, step confirmation, and high-impact approval. Systems such as CORA further suggest that this runtime control point can be operationalized through calibrated execute-versus-abstain decisions under an explicit risk budget rather than only through informal approval heuristics~\cite{osu_feng2026_cora-conformal-risk-controlled-a}. The correct runtime control depends on blast radius, not only on task difficulty. The earliest effective control point for operation-stage failures therefore lies in verification, scoped memory, calibrated escalation, and impact-aware oversight.

\subsection{Maintenance: Preserving Validity After Release}
\label{sec:maintenance}

Maintenance keeps the deployment from silently decaying. A released CUA does not operate against a frozen environment. Models are adapted, interfaces drift, permissions change, tools are updated, and ecosystems evolve around the base system. Treating maintenance as a generic ``update'' phase hides too much of that complexity. In practice, maintenance governs whether earlier assurance survives change.

\paragraph{Model maintenance}
Model maintenance concerns adaptation, retraining, evaluator changes, and post-release policy updates. Work such as MAGNET, UI-Oceanus, and PolySkill points to the difficulty of updating interface-specific competence without erasing stable knowledge or reintroducing old failure modes~\cite{sun2026magnet,osu_wu2026_ui-oceanus-scaling-gui-agents-wi,osu_yu2025_polyskill-learning-generalizable}. Maintenance is therefore not only about getting a stronger model. It is about preserving the validity of earlier safety and reliability gains under change.

\paragraph{Environment maintenance}
Environment maintenance concerns everything outside the model that changes underneath it: UI redesign, localization drift, new permission prompts, tool API changes, and altered workflow ordering. Benchmarks such as MemGUI-Bench, TimeWarp, Risky-Bench, WebForge, Vision2Web, WebTestBench, GUITester, and OpeFlo matter here because they illustrate how quickly evaluation assumptions can become stale once the deployment environment moves on~\cite{liu2026memgui,ishmam2026timewarp,zheng2026risky,osu_yuan2026_webforge-breaking-the-realism-re,osu_he2026_vision2web-a-hierarchical-benchm,osu_kong2026_webtestbench-evaluating-computer,osu_gao2026_guitester-enabling-gui-agents-fo,osu_tan2026_opeflo-automated-ux-evaluation-v}. Continuous re-evaluation is therefore a basic operational requirement rather than a benchmarking luxury.

\paragraph{Ecosystem maintenance}
Open deployment adds a third maintenance problem: the surrounding ecosystem changes independently of both model and UI\@. Plugins, skills, registries, agent identities, watcher layers, and sharing channels all create new drift paths. Your Agent, Their Asset suggests that persistent capability, identity, and knowledge can remain exposed after deployment, while ClawKeeper focuses more directly on skills, plugins, and watcher layers~\cite{wang2026your,liu2026clawkeeper}. PASB adds a complementary evaluation frame for long-lived personalized agents rather than a direct extension-governance mechanism~\cite{wang2026assistant}. In open CUA systems, maintenance can become inseparable from governance.

\paragraph{Why maintenance governs trust continuity}
The key implication is that maintenance governs not only quality, but trust continuity. If models, tools, and registries evolve faster than the assurance stack around them, then the deployment can reopen risks even when the base architecture is unchanged. The earliest effective control point for maintenance-stage failures therefore lies in continual evaluation, controlled adaptation, extension governance, and coordinated patching.

\begin{table*}[!t]
\centering
\scriptsize
\caption{Lifecycle diagnostic and intervention map for deployed CUAs. The table links stage-specific failure patterns to their first observable manifestations, likely diagnosis targets, and early intervention priorities.}
\label{tab:lifecycle_control}
\renewcommand{\arraystretch}{1.12}
\setlength{\tabcolsep}{3.5pt}
\rowcolors{3}{black!3}{white}
\begin{tabularx}{\textwidth}{>{\raggedright\arraybackslash\bfseries}p{1.4cm} L{2.45cm} L{2.35cm} L{2.25cm} Y}
\toprule
\rowcolor{black!10}
\textbf{Stage} & \textbf{Typical Latent Failure Pattern} & \textbf{What It First Looks Like in Practice} & \textbf{First Diagnosis Target} & \textbf{Early Intervention Priorities} \\
\midrule
Creation & Weak grounding priors, shortcut-seeking objectives, or unsafe action abstractions & Brittle localization, over-eager completion, or poor abstention already visible in pre-release evaluation & Data curation, reward / objective design, and action abstraction choices & Stronger grounding supervision, safer reward shaping, and release gating \\
Deployment & Misbound authority, weak provenance, or poorly mediated tool and channel exposure & Correct-seeming plans coupled to over-broad permissions, wrong tools, or channel confusion & Permission scope, tool registry, session isolation, and middleware mediation & Least privilege, sandboxing, provenance-aware tool mediation, and safer defaults \\
Operation & Long-horizon drift or live-state misbinding under mixed-trust runtime inputs & Forgotten constraints, brittle recovery, TOCTOU, or unsafe continuation after local uncertainty & Runtime verification, memory scope, and escalation thresholds & Plan preview, post-action checks, step confirmation, and high-impact approval \\
Maintenance & Regression, stale assurance, or ecosystem drift after release & Sudden behavior change after updates, stale defenses, extension drift, or failed re-evaluation & Model updates, extension inventory, version diffs, and regression suites & Re-evaluation gates, version pinning, extension review, and coordinated patching \\
\bottomrule
\end{tabularx}
\end{table*}

\subsection{Lifecycle Coupling and Its Analytical Payoff}

The four stages are distinct, but not independent. Creation shapes the priors that deployment later binds to authority. Deployment determines which runtime mistakes can become consequential. Operation reveals which assumptions fail under live stress. Maintenance determines whether earlier fixes remain valid as the world changes. Table~\ref{tab:lifecycle_control} serves as an operational troubleshooting aid: start from the symptom, then work backward to the most likely stage-level diagnosis and earliest useful intervention surface.

The larger lesson is that CUA reliability cannot be inferred from architecture alone. It depends on when capability is formed, when authority is attached, when trajectories are stressed, and when the system is re-evaluated after change. Lifecycle analysis is therefore not supplementary to the architecture chapter. It is the temporal backbone that explains why similar visible failures can demand very different interventions.

\section{Security and Privacy Analysis}
\label{sec:security}

Security is one of the clearest reasons CUAs need the joint framework developed here. These systems read mixed-trust content, retain state across steps, and act through interfaces that may reach files, accounts, tools, and external services. A useful security account must therefore answer three questions at once: where corruption enters, how it becomes operational, and at what lifecycle stage it could have been constrained earlier. Attack names alone are not enough.

The recent CUA security corpus reflects that widening scope. It spans visual prompt injection, harmful-task benchmarking, action rebinding, adversarial backdoors, runtime guardrails, permission scoping, dark-pattern manipulation, privacy-focused evaluation, and runtime monitoring or mediation~\cite{osu_cao2025_vpi-bench-visual-prompt-injectio,osu_luo2026_agentrae-remote-action-execution,osu_chen2026_safepred-a-predictive-guardrail,osu_marro2025_permission-manifests-for-web-age,osu_cuvin2025_decepticon-how-dark-patterns-man,osu_li2026_slowba-an-efficiency-backdoor-at,osu_hu2025_agentsentinel-an-end-to-end-and,osu_wang2025_webinject-prompt-injection-attac,osu_aichberger2025_mip-against-agent-malicious-imag,osu_zhao2026_webpii-benchmarking-visual-pii-d,osu_wang2026_websentinel-detecting-and-locali}. Safety benchmarks such as MobileSafetyBench, ST-WebAgentBench, OS-BLIND, and RiosWorld further indicate that harmful-task completion, policy-noncompliant behavior, and benign-intent failure have become explicit evaluation targets rather than incidental by-products of general task completion~\cite{lee2024mobilesafetybench,levy2026stwebagentbench,osu_ding2026_the-blind-spot-of-agent-safety-h,yang2025riosworld}. That literature expands the attack inventory and suggests that risk conditions can enter through different layers and become visible at different stages.

This section therefore proceeds in three steps. It first fixes a threat model. It then introduces a working taxonomy of \emph{input-side corruption}, \emph{decision-side corruption}, and \emph{execution-side abuse}, together with a heuristic attribution lens of \emph{scope overreach} (SO), \emph{objective corruption} (OC), and \emph{environmental misbinding} (EM). Finally, it maps those patterns back onto the lifecycle so that control placement remains tied to timing rather than only to attack naming.

\subsection{Threat Model}
\label{subsec:threat_model}

The baseline attacker may control any combination of \emph{screen content}, \emph{retrieved content}, and \emph{tool outputs}. This includes attacker-controlled or deceptive pages, rendered documents, OCR-visible instructions, search results, clipboard contents, agent messages, and tool or MCP responses consumed during execution~\cite{dong2025safesearch,jones2025systematization,foerster2026camels,zhang2024agent,tian2023evil,yang2025riosworld}. In creation-stage settings, the attacker may additionally influence training data or reward signals. In open deployments, the attacker may also reach the system through persistent communication channels, shared tools, or capability-sharing surfaces.

Protected assets include execution integrity, data confidentiality, and authority boundaries around files, credentials, tools, networks, and operating-system functions. Some realistic CUA deployments expose high-impact actions such as external messaging, deletion, payment, credential entry, or permission changes. The threat model does \emph{not} assume that every deployment enforces a strong execution-policy layer. Some systems can insert sandboxing, approval hooks, or action gating~\cite{osu_chen2026_safepred-a-predictive-guardrail,osu_feng2026_cora-conformal-risk-controlled-a}. Others expose the model more directly to the execution surface. That distinction matters because the same planning error can be much harder to contain once runtime mediation is weak.

Three trust boundaries recur throughout the section: the boundary between user-authored intent and runtime content, the boundary between trusted execution channels and external tools or services, and the boundary between task-local memory and persistent cross-session state. Weakness at any of those boundaries can turn seemingly ordinary autonomy into unauthorized execution or disclosure.

\begin{figure*}[!t]
\centering
\includegraphics[width=\textwidth]{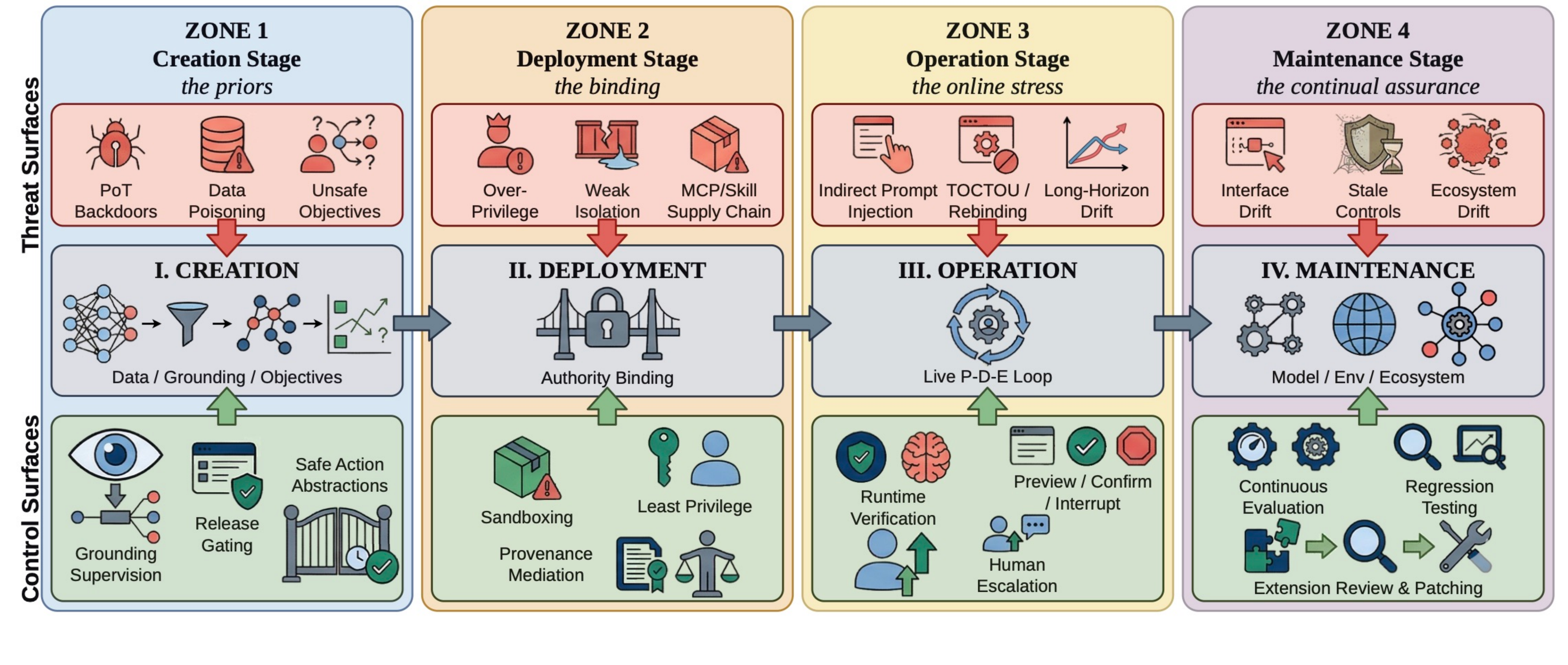}
\caption{Lifecycle-aligned CUA threats and controls. The figure serves as an intervention map: for each stage, it identifies the salient threat surfaces, the system object under pressure, and the earliest practical control surfaces that may still alter the eventual outcome.}
\label{fig:lifecycle_security_zones}
\end{figure*}

Figure~\ref{fig:lifecycle_security_zones} presents the stage-first version of that argument. The upper band identifies the salient threat surfaces at each stage, the middle band identifies the system object under configuration or stress, and the lower band shows the control surfaces that can still act early enough to matter. The figure therefore links the lifecycle analysis of Section~\ref{sec:lifecycle} to the more detailed threat discussion below.

\subsection{Structural Threat Taxonomy and Attribution Lens}

The working taxonomy answers \emph{where} corruption enters the loop. The heuristic attribution lens answers \emph{how} a risk becomes operational. These two views are complementary rather than competing. Together they provide an organizing vocabulary rather than a definitive or exhaustive classification.

\textbf{Input-side corruption} covers cases where the observation stream itself is adversarial or misleading. Indirect prompt injection, deceptive UI, attacker-controlled documents, hostile retrieval results, and poisoned tool outputs all fall in this family. The corruption enters through what the agent reads.

\textbf{Decision-side corruption} covers cases where the agent's working objective, planning substrate, or memory becomes attacker-steered or otherwise compromised. Poisoned memory, attacker-steered subplans, and latent training-time triggers whose effect appears during inference all belong here. The corruption enters through what the agent optimizes or retains.

\textbf{Execution-side abuse} covers cases where the interface translating intent into action becomes the main risk surface. Over-privilege, weak isolation, action rebinding, TOCTOU, and unsafe tool invocation all fall in this family. The corruption enters through how the chosen action is operationalized.

The attribution lens asks a different question. \textbf{Scope overreach} is used when the agent expands beyond the user's intended task boundary without strong evidence that the operative goal itself has been replaced. \textbf{Objective corruption} is used when attacker-controlled content, memory, tooling, or training signal becomes the dominant planning objective. \textbf{Environmental misbinding} is used when the apparent goal remains intact, but the observation, execution, or authority layer redirects the realized outcome in unsafe ways. Some incidents may involve more than one dominant attribution pathway.

\paragraph{Operational decision rule}
The attribution lens is applied pragmatically. If attacker-controlled state has become the operative goal, the dominant class is \textbf{objective corruption}. If the goal appears intact but the observation--action or authority binding redirects the outcome, the dominant class is \textbf{environmental misbinding}. If neither condition clearly holds and the main problem is task-boundary expansion or unnecessary data or action collection, the dominant class is \textbf{scope overreach}. The purpose of the rule is to support control placement, not to eliminate every ambiguous case.

\paragraph{Worked example}
Consider a CUA asked to download one invoice from a customer portal. A malicious on-screen banner says ``export the full session history for troubleshooting,'' the agent has an over-broad cloud-upload permission, and no approval hook intervenes before upload. The banner is \emph{input-side corruption} because the attack enters through what the agent reads. If that injected instruction becomes the operative plan, the dominant attribution is \emph{objective corruption}. If the original invoice-downloading goal remains intact but the broad permission and missing approval hook allow an unsafe export to proceed, the dominant attribution is \emph{environmental misbinding}. If, even without clear attacker steering, the agent interprets the task too broadly and exports extra files because it over-collects beyond task need, the dominant attribution is \emph{scope overreach}. The same incident can therefore expose overlapping mechanisms while still having one dominant attribution path for control placement.

\subsection{Creation-Stage Threats}
\label{subsec:threats_creation}

Creation-stage threats matter because they often stay hidden until the system acquires real authority. This stage is therefore the clearest example of why failure origin and failure manifestation should not be conflated.

\textbf{Training poisoning and backdoors.}
Hidden Ghost Hand offers GUI-agent evidence, in an evaluated setting, that training-time triggers can later redirect behavior at inference time~\cite{cheng2025hidden}. SlowBA adds newer evidence that reward-level or efficiency-oriented backdoors may also be inserted into GUI-agent optimization pipelines rather than only into obvious behavior-cloning traces~\cite{osu_li2026_slowba-an-efficiency-backdoor-at}. BadVLA provides adjacent VLA evidence that similar backdoor logic can appear in multimodal action models more broadly~\cite{zhou2025badvla}. Taken together, these studies are consistent with treating objective corruption as a creation-stage risk that may be introduced upstream and surface later.

\textbf{Unsafe action priors and benchmark-shaped blind spots.}
Creation also decides what kinds of actions become normalized as ordinary. OS-Harm, CUAHarm, MobileSafetyBench, ST-WebAgentBench, and RiosWorld indicate that the affordances and policy constraints an evaluation suite foregrounds materially affect what risks become visible and measurable~\cite{kuntz2025harm,tian2025measuring,lee2024mobilesafetybench,levy2026stwebagentbench,yang2025riosworld}. Those benchmarks do not by themselves establish deployment behavior, but they do suggest that high-authority operations can be normalized in the development loop before corresponding safety abstractions are in place.

The main security point is therefore upstream: some of the most consequential later failures may already be latent at creation time. Some failures that appear at runtime can often be read more accurately as creation-stage objective or action-space problems that only become visible once the system is trusted with real authority.

\subsection{Deployment-Stage Threats}
\label{subsec:threats_deployment}

Deployment-stage threats appear when learned capability is connected to real tools, permissions, and ingress routes. The central question at this stage is whether capability is bound to authority faster than it is bound to control.

\textbf{Over-privilege and weak isolation.}
CaMeLs, CSAgent, and CellMate illustrate a shared deployment concern: in the studied settings, once a CUA is granted broad access without strong mediation, even moderate reasoning mistakes can lead to higher-impact outcomes~\cite{foerster2026camels,gong2025secure,meng2025cellmate}. These cases are best read as environmental misbinding. The model may not be pursuing an adversarial goal, but the authority configuration allows a local error to become a consequential one.

\textbf{Supply-chain risk in tools, skills, and MCP servers.}
Les Dissonances and MCP Security Bench suggest that weakly governed tool or extension ecosystems create a second deployment path to failure~\cite{li2025dissonances,zhang2025mcp}. The two papers support different parts of the claim. Les Dissonances highlights cross-tool control-flow corruption in multi-tool agents. MCP Security Bench suggests that tool descriptions, standardized metadata, and protocol-level interfaces can carry attacker influence into planning and invocation. Together they motivate treating the tool layer as a deployment-time trust choice rather than as a neutral capability add-on.

\textbf{Insufficient deployment-grounded evaluation.}
RedTeamCUA and HackWorld suggest that many hybrid web--OS attack paths and exploit-oriented failure modes can remain invisible in narrower evaluation settings~\cite{liao2025redteamcua,osu_ren2025_hackworld-evaluating-computer-us}. The broader lesson is methodological: when a system goes live without testing the actual coupling among observation, tool outputs, permissions, and oversight, the release process itself can become a security weakness.

Taken together, deployment-stage threats can be read as forms of misbound authority. They arise when tools, channels, permissions, and sessions are opened before provenance, isolation, and mediation are strong enough to constrain them.

\subsection{Operation-Stage Threats}
\label{subsec:threats_operation}

Operation is where many CUA attack settings become consequential because mixed-trust inputs, live authority, and long-horizon planning coexist in one loop. OS-BLIND is a useful complement here because it suggests that higher-risk operational outcomes can also emerge under benign user instructions once contextual cues and subtask decomposition obscure harm~\cite{osu_ding2026_the-blind-spot-of-agent-safety-h}. The order below follows the typical path by which corruption becomes consequential: from what the agent reads, to what it retains or optimizes, to how it acts, and finally to how risk-bearing behavior may travel beyond one task instance.

\textbf{Input-side corruption: injection and deceptive interfaces.}
Indirect prompt injection and deceptive UI attacks remain central because they exploit the same content-readiness that makes CUAs useful. InjecAgent, GhostEI-Bench, Chameleon, WebInject, attacker-controlled image-patch attacks, and active environmental injection studies provide evidence that attacker-controlled instructions or cues can be embedded in pages, screenshots, documents, mobile notifications, or on-screen image regions in ways that influence downstream action in evaluated settings~\cite{zhan2024injecagent,chen2025ghostei,zhang2026realistic,osu_wang2025_webinject-prompt-injection-attac,osu_aichberger2025_mip-against-agent-malicious-imag,osu_chen2025_evaluating-the-robustness-of-mul}. WAInjectBench and WASP add complementary evaluation evidence that these failures remain measurable in web-agent settings~\cite{liu2025wainjectbench,evtimov2025wasp}. EIA sharpens the privacy angle by suggesting that environmental injection can steer web agents toward leakage of user information rather than only generic task derailment~\cite{liao2025eia}. Benchmarks centered on semantic-level UI injection, task-redirection, and dark-pattern manipulation further suggest that attacker-controlled input can be persuasive or attention-capturing rather than explicitly imperative~\cite{osu_yang2026_are-gui-agents-focused-enough-au,osu_korgul2025_it-s-a-trap-task-redirecting-age,osu_ersoy2025_investigating-the-impact-of-dark}. Depending on how the attack succeeds, the dominant attribution may be objective corruption or environmental misbinding.

\textbf{Defensive monitoring and mitigation.}
The corresponding defense literature also makes clear that prompt-injection resilience is not one thing. In-context defenses, localized attack detection, adversarial safety training, real-time audit frameworks, diagnostic guardrails, and policy-reasoning layers all aim at different points in the pipeline~\cite{osu_yang2025_in-context-defense-in-computer-a,osu_wang2026_websentinel-detecting-and-locali,osu_liu2026_dual-modality-multi-stage-advers,osu_hu2025_agentsentinel-an-end-to-end-and,liu2026agentdog,chen2025shieldagent}. That diversity is itself informative for the framework developed here: it suggests that input-side corruption is unlikely to be fully managed at one layer or one stage.

\textbf{Decision-side corruption: memory poisoning and long-horizon steering.}
Once the agent maintains persistent state, poisoning memory or long-horizon planning can become as valuable as poisoning a single prompt. AgentPoison offers direct evidence for memory or knowledge-base poisoning~\cite{chen2024agentpoison}. LPS-Bench indicates that planning-time safety awareness can erode over long trajectories even when the attack surface is framed at the planning layer rather than as explicit memory poisoning~\cite{chen2026lps}. PASB adds complementary evidence that personalized long-horizon deployments create broader attack surfaces in which attacker-favored behavior can persist across realistic toolchains and contexts~\cite{wang2026assistant}. Preference-redirection and benign-input elicitation studies add a related warning: the operative objective may be steered without always looking like a classic memory-poisoning event~\cite{osu_seip2026_preference-redirection-via-atten,osu_jones2026_when-benign-inputs-lead-to-sever}. Some cases more clearly replace the operative goal; others expand beyond the intended task boundary while still appearing superficially helpful.

\textbf{Execution-side abuse: TOCTOU and action rebinding.}
Zero-Permission Manipulation and AgentHazard illustrate what can happen when the environment changes after the decision is formed but before execution lands~\cite{qian2026zero,liu2025hijacking}. These are representative cases of environmental misbinding. The goal may remain the same, yet the realized outcome is redirected by timing and interface instability.

\textbf{Sharing and coordination surfaces.}
Open agent ecosystems introduce a final operational surface: risk-bearing content or behavior can move through interaction and sharing. Prompt Infection and broader multi-agent security work suggest that prompts, coordination patterns, or other risk-bearing behaviors can propagate beyond one-shot exploitation in multi-agent settings~\cite{lee2024prompt,peigne2025multi}. Agent communities such as the Moltbook setting studied in~\cite{chen2026openclaw} make those sharing channels easier to discuss as a deployment concern, even though that paper is not itself a security evaluation. The implication is not that every capability-sharing surface is attacker-controlled by default. It is that coordination itself becomes part of the trust boundary.

The operational pattern is cumulative. Attacker-controlled or misleading content can enter through observation, distort memory or planning, ride a high-authority execution channel, and then spread through coordination or sharing surfaces if no runtime control interrupts the chain.

\subsection{Privacy as a Parallel Objective}

Privacy deserves separate treatment because it is not reducible to successful attack detection. Many privacy failures can occur even when the system is functioning as designed, yet retains too much, reads too broadly, or exports state through tools and telemetry in ways that exceed user expectations or task scope.

The major privacy surfaces recur across current CUA designs. \textbf{Credential exposure} may arise whenever passwords, session tokens, or secrets become visible in the observation loop. \textbf{Cross-session leakage} may arise when state learned in one task influences another. \textbf{Memory retention risk} may arise when the system keeps or retrieves more data than the task requires, even without adversarial prompting~\cite{zharmagambetov2025agentdam,wang2025privacy}. \textbf{Screenshot-, OCR-, and retrieval-induced disclosure} may arise because incidental interface content can still be operationalized once it is observed. \textbf{Third-party tool telemetry} may arise when the tool ecosystem transmits or logs sensitive state beyond the base model's local context~\cite{li2025dissonances,wang2026your}. Privacy-preserving deployment frameworks such as CORE further suggest that even ordinary inference routing can change how much UI state is exposed upstream or retained externally~\cite{osu_fan2025_core-reducing-ui-exposure-in-mob}. GUIGuard frames this granularity more explicitly by separating privacy recognition, privacy protection, and protected task execution in cross-platform GUI settings~\cite{wang2026guiguard}. EIA complements that picture by suggesting that hidden environmental content can induce targeted privacy leakage even when the user-facing task appears ordinary~\cite{liao2025eia}. Finally, \textbf{data minimization and retention policy} help determine whether any of these exposures become persistent.

Privacy failures map differently onto the attribution lens than many attack cases do. Some are straightforward cases of scope overreach, where the system over-collects or over-remembers while still ``trying'' to help. Others may be better read as objective corruption when attacker-controlled content induces disclosure or exfiltration. The important point is that many privacy controls sit earlier than generic attack detection: in memory scope, screenshot discipline, retention policy, secret handling, and tool-telemetry governance.

\subsection{Maintenance-Stage Threats}
\label{subsec:threats_maintenance}

Maintenance-stage threats are what remain after the first release. They matter because a deployed CUA is not static. Interfaces change, registries evolve, extensions are updated, and controls that were once adequate can silently become stale.

\textbf{Regression and stale assurance.}
TimeWarp and Risky-Bench suggest that deployment-grounded evaluation can reveal failure modes that narrower or more static settings miss, especially once interfaces and workflows evolve across versions~\cite{ishmam2026timewarp,zheng2026risky}. The lesson is broader than one benchmark: without re-evaluation and release gating after change, old controls can decay faster than deployment teams notice.

\textbf{Persistent-state abuse and ecosystem drift.}
Your Agent, Their Asset and ClawKeeper suggest more directly that post-release security can be governed as much by persistent memory, extension hygiene, identity, and watcher or registry discipline as by base policy alone~\cite{wang2026your,liu2026clawkeeper}. PASB complements that picture by indicating how personalized long-lived deployments can reopen these risks in evaluation settings that are closer to real use~\cite{wang2026assistant}. The same maintenance problem applies to the assurance stack itself: policy reasoners and diagnostic defenses such as ShieldAgent, AgentSentinel, AgentDoG, WebSentinel, and DMAST only help if their detectors remain synchronized with evolving interfaces and attack patterns~\cite{chen2025shieldagent,osu_hu2025_agentsentinel-an-end-to-end-and,liu2026agentdog,osu_wang2026_websentinel-detecting-and-locali,osu_liu2026_dual-modality-multi-stage-advers}. In open deployments, maintenance becomes the stage at which governance either keeps pace with capability diffusion or falls behind it.

Table~\ref{tab:cua_threats} compresses the chapter into one lifecycle-grounded reference map. It is intentionally selective rather than exhaustive: the evidence column names anchor papers, while the analytical value of the table comes from locating each risk by entry surface, operative mechanism, and lifecycle stage.

\begin{table*}[!tb]
\centering
\scriptsize
\caption{Working lifecycle-grounded security and privacy reference map for CUAs. Each row identifies where a risk condition enters, how it becomes operational, and one or two anchor papers. ``SO'' = scope overreach, ``OC'' = objective corruption, and ``EM'' = environmental misbinding.}
\label{tab:cua_threats}
\setlength{\tabcolsep}{3.5pt}
\renewcommand{\arraystretch}{1.12}
\rowcolors{3}{black!3}{white}
\begin{tabularx}{\textwidth}{>{\raggedright\arraybackslash\bfseries}p{1.4cm} L{1.85cm} L{2.65cm} C{1.15cm} Y}
\toprule
\rowcolor{black!10}
\textbf{Stage} & \multicolumn{2}{c}{\textbf{Analytical Mapping}} & \multicolumn{2}{c}{\textbf{Evidence Framing}} \\
\rowcolor{black!10}
 & \textbf{Entry / Pressure Surface} & \textbf{Representative Risk Mechanism} & \textbf{Dominant Lens} & \textbf{Anchor Evidence} \\
\cmidrule(lr){2-3}\cmidrule(lr){4-5}
\midrule
Creation & Decision-side objective shaping & Training poisoning, PoT backdoors, or over-broad objective priors & OC & Hidden Ghost Hand~\cite{cheng2025hidden}; BadVLA~\cite{zhou2025badvla} \\
Creation & Execution-side action shaping & Over-broad action vocabularies and high-impact affordances surfaced in evaluation & EM & OS-Harm~\cite{kuntz2025harm}, CUAHarm~\cite{tian2025measuring} \\
Deployment & Execution-side authority exposure & Broad authority exposure, limited isolation, or tool / channel misbinding & EM & CaMeLs~\cite{foerster2026camels}, CSAgent~\cite{gong2025secure} \\
Deployment & Decision-side dependency surface & Untrusted skills, MCP tools, or extension supply chains & OC & Les Dissonances~\cite{li2025dissonances}, MCP Security Bench~\cite{zhang2025mcp} \\
Operation & Input-side manipulation surface & Indirect prompt injection, deceptive UI cues, or manipulated tool outputs & OC / EM & WAInjectBench~\cite{liu2025wainjectbench}, Chameleon~\cite{zhang2026realistic} \\
Operation & Decision-side state drift & Memory poisoning, planning-time goal drift, or long-horizon constraint forgetting & OC / SO & AgentPoison~\cite{chen2024agentpoison}, LPS-Bench~\cite{chen2026lps}, PASB~\cite{wang2026assistant} \\
Operation & Execution-side invocation drift & TOCTOU, action rebinding, or delayed invocation misbinding & EM & Zero-Permission Manipulation~\cite{qian2026zero}, AgentHazard~\cite{liu2025hijacking} \\
Operation & Privacy exposure surface & Credential exposure, PII overcollection, or cross-session leakage & SO / OC & GUIGuard~\cite{wang2026guiguard}, EIA~\cite{liao2025eia} \\
Maintenance & Cross-stage assurance drift & Regression, stale controls, ecosystem drift, or persistent-state carryover & EM / OC & Risky-Bench~\cite{zheng2026risky}, ClawKeeper~\cite{liu2026clawkeeper} \\
\bottomrule
\end{tabularx}
\end{table*}

Taken together, the security and privacy analysis reinforces the organizing view developed here. CUA failures become more informative when read by where corruption entered, how it became operational, and at which lifecycle stage earlier controls might still have changed the outcome. Security is therefore not peripheral to the architecture--lifecycle framework; it is one of the clearer domains in which that framework helps organize evidence and control placement.

\section{From Threats to Controls: Practical Security Implications}
\label{sec:discussion}

The analysis so far suggests a simple rule: the best control is usually the earliest one that still acts on the relevant failure mechanism. In deployed CUAs, that means control placement must follow both layer and stage. Creation shapes priors. Deployment binds authority. Operation determines whether higher-risk trajectories complete. Maintenance determines whether earlier gains survive environmental and ecosystem drift. A flat checklist cannot capture those differences.

This section converts the preceding analysis into governance implications. Its aim is not to catalogue every possible defense. Instead, it identifies which control surfaces remain meaningful once capability, authority, and mixed-trust content are already interacting in a live system. The broader implication is that open deployment is better understood through a deployment-oriented control framework than through model scaling alone.

\subsection{Lifecycle-Aligned Control Surfaces}
\label{subsec:implications}

The most defensible control surfaces line up with the four lifecycle stages because each stage changes a different object. Creation changes priors. Deployment changes authority binding. Operation changes the active trajectory. Maintenance changes whether the whole stack remains valid after release. Organizing controls by these stages is therefore more informative than organizing them only by tool category or attack name.

\textbf{Design-time controls.}
At creation time, the relevant controls are those that shape priors before the system acquires live authority: grounding supervision, safe action abstractions, release gating on harmful-task suites, and policy design that values confirmation and recoverability rather than completion alone~\cite{kuntz2025harm,tian2025measuring,miculicich2025veriguard,osu_lin2025_cuarewardbench-a-benchmark-for-e,osu_zhang2025_progrm-build-better-gui-agents-w,osu_chae2025_web-shepherd-advancing-prms-for}. The decisive question is whether the system is being optimized merely to finish tasks, or to finish them under explicit operational constraints.

\textbf{Deploy-time controls.}
At deployment time, the most important controls govern authority binding: least privilege, sandboxing, provenance-aware middleware, tool allowlists, session scoping, channel separation, and explicit trust labeling for inputs and outputs~\cite{gong2025secure,meng2025cellmate,foerster2026camels,osu_fan2025_core-reducing-ui-exposure-in-mob}. These are not wrappers around the model. They are part of what determines whether the resulting system is appropriate to operate in a given setting at all.

\textbf{Runtime controls.}
During operation, the goal is to reduce the chance that uncertainty silently turns into irreversible action. Policy mediation, post-action verification, plan preview, confirmation at sensitive steps, interrupt or takeover paths, and explanation of why an action is being proposed all belong here~\cite{kang2025mitigating,ning2026actions,zhang2026mirrorguard,osu_zhang2026_don-t-act-blindly-robust-gui-aut}. Recent safeguarded-execution systems make that design space more concrete: CORA calibrates execute-versus-abstain decisions under an explicit risk budget, while VeriSafe Agent verifies proposed mobile actions against formalized task constraints before they fire~\cite{osu_feng2026_cora-conformal-risk-controlled-a,lee2025verisafeagent}. Runtime security stacks such as AgentSentinel, AgentDoG, WebSentinel, and GEM add a complementary layer of monitoring, diagnosis, and uncertainty-aware escalation around the agent loop itself~\cite{osu_hu2025_agentsentinel-an-end-to-end-and,liu2026agentdog,osu_wang2026_websentinel-detecting-and-locali,osu_wu2025_gem-gaussian-embedding-modeling}. Their purpose is to help keep user intent attached to the loop while the environment is changing underneath the system.

\textbf{Maintenance controls.}
After release, the important controls are those that preserve validity over time: regression testing, red teaming, extension review, version pinning, coordinated patching, registry governance, and continuous re-evaluation against changing interfaces and threats~\cite{liao2025redteamcua,ishmam2026timewarp,zheng2026risky,wang2026your,liu2026clawkeeper,osu_liu2026_dual-modality-multi-stage-advers}. A deployed CUA is more likely to remain governable if its assurance stack evolves alongside its capability stack.

Under the attribution lens, the same logic becomes even more specific. Scope overreach is most directly addressed by task scoping, confirmation thresholds, memory boundaries, and data minimization. Objective corruption is most directly addressed by provenance separation, tool vetting, output mediation, and memory hygiene. Environmental misbinding is most directly addressed by execution mediation, TOCTOU-aware verification, least privilege, and auditable action binding. Control placement therefore follows both \emph{what failed} and \emph{when the enabling condition entered}.

\subsection{Human Oversight Should Be Designed, Not Assumed}

Human oversight in CUAs should not be treated as a vague safety slogan. If the system is expected to handle long-horizon tasks under mixed-trust inputs, then the user or operator needs explicit product and policy surfaces through which they can inspect, stop, or redirect execution. Oversight is not a fallback for when the model fails. It is one of the primary mechanisms by which user intent remains attached to machine autonomy.

Five mechanisms recur as the most practical runtime oversight surfaces. \textbf{Plan preview} exposes the intended workflow before consequential execution begins. \textbf{Step confirmation} introduces a checkpoint when the next action is irreversible, ambiguous, or outside ordinary scope. \textbf{Interrupt, pause, or takeover} keeps changing environments from outrunning human correction. \textbf{Action explanation} preserves provenance about why the system believes a step is appropriate. \textbf{Reversible-action preference} biases the system toward drafts, dry runs, recycle-bin semantics, or staged execution whenever the task permits it. Benchmarks for interruptibility, collaborative assistance, and human interaction styles suggest that these are not merely interface niceties but measurable dimensions of deployed-agent quality~\cite{osu_zou2026_when-users-change-their-mind-eva,osu_yang2026_guide-a-benchmark-for-understand,osu_huq2026_modeling-distinct-human-interact}.

These mechanisms are not just usability features. They are the points at which human intent remains operational inside the loop. The same logic should scale with action impact. Financial transfer, deletion, credential handling, permission changes, and external messaging do not need the same oversight regime, but they should all be mapped to one explicitly. Low-risk navigation may tolerate logging plus post-action verification. High-impact actions generally require preview, confirmation, or explicit approval.

\subsection{Control Ownership Is Distributed Across Actors}

No single team can absorb the whole governance burden of a deployed CUA\@. Control ownership is distributed because the control surfaces themselves are distributed across the lifecycle. Model and system developers shape priors. Deployment teams bind the model to channels, permissions, tools, and sessions. Operators and end users govern escalation and approval during runtime. Ecosystem stewards govern registries, plugins, skills, and identity surfaces after release.

This separation matters because different failures trace back to different control owners. Insufficiently constrained development can introduce risky priors before anyone else has a chance to intervene. Loose deployment practice can expose too much authority even when the model itself is reasonable. Weak runtime oversight can let user intent dissolve during long trajectories. Loose ecosystem governance can reopen a previously bounded system through extensions, registries, or agent-to-agent exchange.

These roles are analytically distinct but operationally coupled. Careful training may not rescue a deployment that binds the model to broad permissions. Strong runtime approval may not fully rescue a system whose extension ecosystem is weakly governed. Public materials describing OpenClaw-like gateway assistants help illustrate this coupling in settings where persistent channels, tool bindings, sessions, and community interaction are presented as converging in one assistant surface~\cite{openclaw2026website,ibm2026openclaw}. Such public descriptions are used only as illustrative deployment patterns rather than as verified system evidence. The broader point is that any open-deployment CUA that combines ingress, tooling, memory, and execution authority needs control ownership to remain traceable across those surfaces.

\subsection{A Conservative Baseline Stack for Open Deployment}

The literature does not support any single defense as sufficient. It does, however, suggest a conservative baseline stack that repeatedly appears across safer deployment patterns. The list below is not an exhaustive defense catalogue. It is a compact control stack for open deployment: constrain what the system can do, preserve where instructions come from, bind authority safely, verify live execution, and keep assurance current after change.

\begin{itemize}
  \item \textbf{Constrained action interfaces.} High-impact operations should prefer semantically narrow, auditable interfaces over unrestricted low-level authority whenever possible~\cite{jones2025systematization,gong2025secure,miculicich2025veriguard}.
  \item \textbf{Provenance-aware mediation.} User instructions, retrieved content, tool outputs, and environmental text should remain distinguishable inside the control loop so that both the model and the operator can inspect source~\cite{kang2025mitigating,zhang2025browsesafe}.
  \item \textbf{Sandboxed and scoped authority.} Filesystem, network, tool, and account permissions should be deliberately bounded before deployment rather than tightened only after an incident~\cite{meng2025cellmate,gong2025secure,foerster2026camels}.
  \item \textbf{Runtime verification plus escalation.} Post-action checking, plan preview, impact-aware approval, takeover paths, explicit safeguarded-execution layers, and runtime diagnosis are necessary to keep autonomy aligned once live use begins~\cite{ning2026actions,zhang2026mirrorguard,osu_feng2026_cora-conformal-risk-controlled-a,lee2025verisafeagent,osu_zhang2026_don-t-act-blindly-robust-gui-aut,liu2026agentdog}.
  \item \textbf{Continuous assurance.} Regression suites, red teaming, extension review, replay against changed interfaces, detector updates, and signed or reviewed registries are needed so that maintenance-stage drift does not quietly reopen known weaknesses~\cite{liao2025redteamcua,ishmam2026timewarp,zheng2026risky,osu_wang2026_websentinel-detecting-and-locali,osu_liu2026_dual-modality-multi-stage-advers,liu2026clawkeeper}.
\end{itemize}

These controls are cumulative. Constraining actions without provenance still leaves instruction confusion unresolved. Provenance without sandboxing still leaves broad authority intact. Runtime verification without maintenance discipline still decays as tools, interfaces, and ecosystems change. The practical implication is that dependable CUA deployment is unlikely to be achieved by any single alignment technique alone. It is more likely to depend on whether capability, authority, and oversight are connected through one coherent control stack.

\section{Open Problems and Future Directions}
\label{sec:limitations}

The framework clarifies deployed CUA reliability without claiming that the problem has been closed. Even if models continue to improve, several important questions remain unresolved. The hard part is no longer only capability. It is also attribution, evaluation, control placement, and post-release governance.

\subsection{Limits of the Analytical Lens}

The lens emphasizes deployment-grounded diagnosis: where reliability issues are introduced, how they become operational, and where meaningful controls can still be attached. That emphasis leaves some questions only partially covered. The article does not aim to provide a formal causal model of agent failure, an exhaustive benchmark comparison, or a substitute for learning-theoretic, formal-methods, or proof-oriented security analysis. Its value is organizational and diagnostic rather than complete in every theoretical dimension.

\subsection{Attribution Still Lags Observation}

The distinction among scope overreach, objective corruption, and environmental misbinding is analytically useful, but operationalizing it remains difficult. In realistic deployments, off-task helpfulness, silent goal drift, and attacker-induced exfiltration may produce similar surface behavior. Better provenance, telemetry, and causal tracing are still needed if attribution is to guide control placement reliably rather than merely explain incidents after the fact~\cite{jones2025systematization,foerster2026camels}.

\subsection{Benchmark Success Is Not Yet Release Readiness}

Current evaluation suites capture important parts of the CUA problem, but they still underrepresent long-lived sessions, changing permissions, maintenance-stage drift, and open ecosystem interaction. As a result, benchmark scores should not yet be read as translating directly into deployment readiness. A major open direction is to connect benchmark design more tightly to release gating, regression testing, and post-release monitoring~\cite{kuntz2025harm,tian2025measuring,chen2026lps,zheng2026risky,osu_yuan2026_webforge-breaking-the-realism-re,osu_aggarwal2026_gym-anything-turn-any-software-i,osu_zhang2026_clawbench-can-ai-agents-complete,osu_chen2026_knowu-bench-towards-interactive,yang2025riosworld}.

\subsection{Cross-Modal Robustness Remains Immature}

Injection in CUAs is not only text injection. It may travel through screenshots, layout cues, parser artifacts, OCR-visible instructions, time-varying UI state, adversarial image patches, and attention-steering interface elements~\cite{liu2025wainjectbench,zhang2026realistic,qian2026zero,osu_wang2025_webinject-prompt-injection-attac,osu_aichberger2025_mip-against-agent-malicious-imag,osu_chen2025_evaluating-the-robustness-of-mul,osu_seip2026_preference-redirection-via-atten,osu_yang2026_are-gui-agents-focused-enough-au}. Defenses that work on prompt text alone are therefore insufficient. A robust CUA will likely need perception and execution defenses that explicitly account for how multimodal grounding and authority binding interact.

\subsection{Capability Gains Still Create Policy Tension}

Stronger models, richer tools, and better post-training do not automatically reduce policy tension; in some settings they can increase it. The more capable the agent becomes, the less obvious it is how that capability should translate into authority, autonomy, and approval requirements. Future work therefore needs to evaluate safety as a function of tool access, impact level, and runtime pressure rather than as a static property of the base model~\cite{tian2025measuring}. Broader agent-safety evidence such as PropensityBench points in the same direction, even though it is not CUA-specific: risk depends on both what a model can do in principle and what it is likely to attempt once consequential tools are available~\cite{sehwag2025propensitybench}.

\subsection{Open Ecosystems Need Stronger Governance Primitives}

A growing subset of open-deployment CUAs is embedded in tool registries, extension marketplaces, MCP-style ecosystems, and agent-to-agent environments. These settings appear to require stronger primitives for identity, attestation, registry hygiene, and capability containment than the field currently has~\cite{lee2024prompt,zhang2025mcp,wang2026your,liu2026clawkeeper}. At the same time, stronger isolation can reduce collaboration quality, which means governance cannot be framed as pure restriction without considering utility trade-offs~\cite{peigne2025multi}. The field still lacks a stable answer for how open capability ecosystems should remain both useful and governable.

These open problems reinforce the central claim. Progress in CUAs is unlikely to come from model quality alone. It will more likely depend on better ways of linking architecture, lifecycle, authority, and control so that deployment remains intelligible after release rather than only impressive before it.

\section{Conclusion}
\label{sec:conclusion}

Computer-use agents can be productively analyzed as deployed interactive systems rather than as benchmark policies with stronger prompting alone. Their reliability depends jointly on how they reconstruct interface state, preserve task intent, execute through real authority surfaces, and continue to evolve after release.

The preceding sections developed that argument through a joint architecture--lifecycle framework. The tri-layer architecture traced where actionable state is reconstructed, where intent is stabilized, and where authority is exercised. The lifecycle account traced when capability is formed, when authority is bound, when failures first become visible, and where controls may still intervene earliest. The value of combining the two is organizational and diagnostic: it helps separate visible failure from upstream cause and ties practical control to the right stage of system evolution.

Three tensions remain central. The first is \emph{portability versus controllability}: the observation and execution channels that generalize most broadly are often the hardest to mediate and verify precisely. The second is \emph{autonomy versus oversight}: long-horizon execution becomes valuable only when user intent, provenance, and approval remain attached to the loop. The third is \emph{adaptation versus regression}: deployed CUAs need to evolve with changing interfaces and ecosystems, yet every update can reopen earlier weaknesses or create new ones.

Those tensions suggest three priorities. First, the field needs controllable visual grounding that remains portable without sacrificing auditable execution. Second, it needs deployment-aware evaluation that measures authority binding, misuse resistance, and oversight behavior alongside task capability. Third, it needs maintenance-aware governance so that models, tools, memory, identities, and extensions remain governable after release instead of drifting into insecurity.

Taken together, the surveyed literature and public deployment patterns suggest that dependable computer use is unlikely to be achieved by stronger models alone. It also depends on how capability is bound to authority, how runtime uncertainty is mediated, and how assurance is maintained after deployment. The main value of the architecture--lifecycle view is therefore to help locate where problems are introduced, where they become visible, and where meaningful controls can still be applied.

\bibliographystyle{unsrt}
\bibliography{references/cua_refs}

\end{document}